\documentclass{article}

\usepackage[eandd, final]{neurips_2026}
\usepackage[pdftex]{graphicx}
\usepackage{float}
\usepackage[utf8]{inputenc} 
\usepackage[T1]{fontenc}    
\usepackage{hyperref}       
\usepackage{url}            
\usepackage{booktabs}       
\usepackage{amsfonts}       
\usepackage{nicefrac}       
\usepackage{microtype}      
\usepackage{xcolor}         

\title{CogScale:\\Scalable Benchmark for Sequence Processing}

\author{%
  Yannis Bendi-Ouis\\
  Mnemosyne\\
  Inria\\
  Bordeaux, France \\
  \texttt{yannis.bendi-ouis@inria.fr} \\
  \and
  Romain de Coudenhove \\
  ENS PSL \\
  École normale supérieure (Ulm) \\
  Paris, France \\
  \texttt{romain.de.coudenhove@ens.psl.eu} \\
  \and
  Xavier Hinaut\\
  Mnemosyne\\
  Inria\\
  Bordeaux, France \\
  \texttt{xavier.hinaut@inria.fr} \\
}

\begin{document}

\newcounter{xaviercounter}
\newcommand{\xav}[1]
{\stepcounter{xaviercounter}
 {\color{blue} #1 (XH\thexaviercounter)}
 }

\newcounter{yanniscounter}
\newcommand{\yannis}[1]
{\stepcounter{yanniscounter}
 {\color{red} #1 (BOY\thexaviercounter)}
 }

\maketitle

\begin{abstract}
  The ability to maintain and manipulate information over time is a fundamental aspect of living beings and Artificial Intelligence.
  While modern models have achieved remarkable success in tasks like natural language processing, evaluating the capacity of novel architectures to process sequential information remains computationally expensive and time-consuming. Testing a new architecture often requires scaling up to massive datasets and models, leading to vast computational costs and slow iteration cycles. In this paper, we propose CogScale, a benchmark of 14 scalable synthetic tasks designed to isolate and evaluate specific cognitive and memory abilities at different parametrizable scales. By providing a standardized, lightweight framework, CogScale allows researchers to rapidly validate architectural innovations before committing to large-scale training. To establish a solid baseline, we evaluate seven distinct architectures: Gated Recurrent Unit (GRU), Long Short-Term Memory (LSTM), xLSTM, Echo State Network (ESN), Mamba, Transformer Decoder, and Transformer Encoder-Decoder. These evaluations are conducted under strict parameter budgets (1k, 10k, and 100k) and across different difficulty levels and scales. Our results show that while classical RNNs and Echo State Networks excel at basic retention within strict parameter budgets, only attention mechanisms and modern state-space models consistently maintain high performance as reasoning complexity and task difficulty scale.
\end{abstract}


\section{Introduction}

The development of foundational models like Large Language Models (LLMs) requires architectures that can handle a diverse set of cognitive skills, ranging from information retention to complex reasoning. Historically, the evaluation of these capabilities has relied on massive Natural Language Processing (NLP) benchmarks, such as Lambada \cite{paperno2016lambada}, WinoGrande \cite{sakaguchi2021winogrande}, PiQA \cite{bisk2020piqa} or HellaSwag \cite{zellers2019hellaswag} and necessitate training on massive datasets like OpenWebText (40GB) \cite{Gokaslan2019OpenWeb} or The Pile (880GB)~\cite{gao2020pile}. While these evaluations are needed to assess the overall performance of large language models, they only demonstrate meaningful distinctions when evaluated on models with massive parameter counts. 
This demands an exorbitant amount of computational resources and time, making  such approaches highly prohibitive for early architectural exploration.
This excludes the analysis of small models and strongly penalizes studies 
proposing novel architectures, as the training requirements make architectural iteration cycles slow and costly.
Beyond the economic cost, the massive energy consumption required by current scaling laws poses a significant environmental challenge \cite{strubell2019energy, patterson2021carbon}, highlighting the need for more sustainable research methodologies \cite{schwartz2020green}.
Recent critiques of the ``bigger-is-better''~\cite{varoquaux2025hype} paradigm also highlight how this scaling view of AI can reinforce unequal access to research by tying scientific progress to expensive computational infrastructure.
It also leaves out researchers with no or limited computing resources and creates a consequent gap between academia (even more in low and middle-income countries) and large companies.
Furthermore, existing datasets often lack scalability; for instance, the TSL benchmark \cite{wang2024tssurvey} is constrained to 69 tasks with fixed sizes and difficulties. 

To resolve this issue, we introduce CogScale, a lightweight synthetic benchmark designed to serve as a cognitive "sanity check" before initiating any massive language model training. We argue that if a new architecture (aiming to scale on large datasets) is fundamentally incapable of resolving basic cognitive and sequential tasks, or fails to approach the baseline performance of a standard Transformer, it is both futile and environmentally costly to attempt to scale it on massive datasets of text. 
Beyond serving as a prerequisite for massive scaling, CogScale is equally valuable for developing small efficient models that must generalize from limited data \cite{varoquaux2025hype}.
CogScale provides a minimal evaluation framework, allowing researchers to rapidly validate their architecture for a fraction of the traditional cost. A major feature of CogScale is the scalability of its tasks. Unlike static benchmarks, CogScale allows researchers to modulate the difficulty and complexity of each task, enabling the assessment of various levels of expertise for the same cognitive skill. This scalability is required for verifying that a model's performance improves effectively as its parameter count increases. Since more complex versions of a task demand greater representational capacity from the model and often require handling larger input and output dimensions, CogScale provides a precise diagnostic tool to ensure that scaling a model's size genuinely translates to better cognitive capabilities.

In this paper, our contributions are twofold. First, we introduce the CogScale framework, which includes 14 scalable synthetic tasks designed to evaluate specific cognitive abilities. Second, we provide a baseline evaluation of seven distinct architectures, including LSTM \cite{hochreiter1997long, gers2000learning}, GRU \cite{cho2014learning}, Transformer Encoder-Decoder (ED) \cite{vaswani2017attention} and Decoder-Only (DO) \cite{radford2018improving}, Mamba \cite{gu2023mamba}, xLSTM \cite{beck2024xlstm} and an atypical but competitive dynamical system via Echo State Networks (ESN) \cite{jaeger2001echo, jaeger2002adaptive}. This experiment is conducted under a rigorous evaluation protocol under strict parameter budgets (1k, 10k, and 100k parameters). It allows us to highlight the strengths and weaknesses of each tested architecture and provides a methodology for ensuring fair comparisons in future architectural research.


\section{Related Work}

Many different benchmarks propose to isolate specific cognitive abilities without the computational cost of training on massive datasets. Notable works include the Long Range Arena (LRA) \cite{tay2020long}, designed to evaluate long context retrieval and spatial reasoning, bAbI \cite{weston2015towards}, which introduces toy tasks for logical reasoning, and the Time Series Library \cite{wang2024tssurvey} for continuous temporal dynamics. However, these existing datasets suffer from structural rigidity, consisting of fixed sequence lengths, static difficulty levels, and thus, a lack of scalability. CogScale distinguishes itself through its modularity. By offering configurable sequence lengths, vocabulary sizes, and difficulty levels, CogScale provides an adaptable evaluation framework that overcomes the rigidity of its predecessors, allowing architectures to be evaluated at different difficulty levels and scales.

In order to establish a comparative baseline, a diverse set of sequential architectures has been evaluated. We include classical Recurrent Neural Networks (RNNs), specifically Long Short-Term Memory (LSTM) and Gated Recurrent Units (GRU). The field's current standard, the Transformer, is also tested, allowing us to contrast the original Encoder-Decoder (ED) with the Decoder-Only (DO) architecture. Furthermore, we evaluate recent innovations in sequence processing, notably State Space Models with Mamba and advanced recurrent architectures such as xLSTM, which offer promising alternatives to standard attention mechanisms.
Finally, we include the Echo State Network (ESN) from the Reservoir Computing paradigm~\cite{lukovsevivcius2009reservoir, yan2024emerging} into our evaluation. While ESNs represent an atypical baseline in contemporary deep learning due to their reliance on a fixed, randomly initialized recurrent reservoir rather than end-to-end backpropagation, they are strong contenders when number of parameters is limited. Including them provides a reference point, allowing us to assess whether the complex modern architectures genuinely yield superior temporal representations or if simple reservoir dynamics suffice for certain cognitive tasks, at a certain difficulty.

\newpage
\section{The CogScale Dataset}

\vspace{-5pt}

\subsection{Synthetic Generation and Unified Evaluation}

\vspace{-5pt}
Because all tasks in CogScale are synthetic, data samples can be generated dynamically during the training process by setting a specific seed to ensure full reproducibility. This methodological choice presents several major advantages for architectural exploration: it requires zero disk storage, bypassing the bottleneck associated with loading massive datasets, and provides a theoretically infinite data distribution. By perpetually generating novel sequences, the framework effectively prevents models from simply memorizing, forcing them to learn the underlying generative rules of the tasks. Furthermore, CogScale uses a unified evaluation metric system. Depending on the task's category, the benchmark evaluates the model on designated prediction timesteps using Mean Squared Error (MSE) for regression, error rate (1-Accuracy) for classification, and label-based error rate (1-Label Accuracy) for multi-label classification. All the source code required to generate the CogScale dataset is publicly available.\footnote{\url{https://anonymous.4open.science/r/CogScale/}}
\vspace{-10pt}

\subsection{Signal Processing and Forecasting}

\subsection{Sinus Forecasting}
\vspace{-25pt}
\noindent
\begin{figure}[htbp] 
    \centering
    \begin{minipage}[c]{0.48\textwidth}
        \centering
        \includegraphics[width=1.00\textwidth]{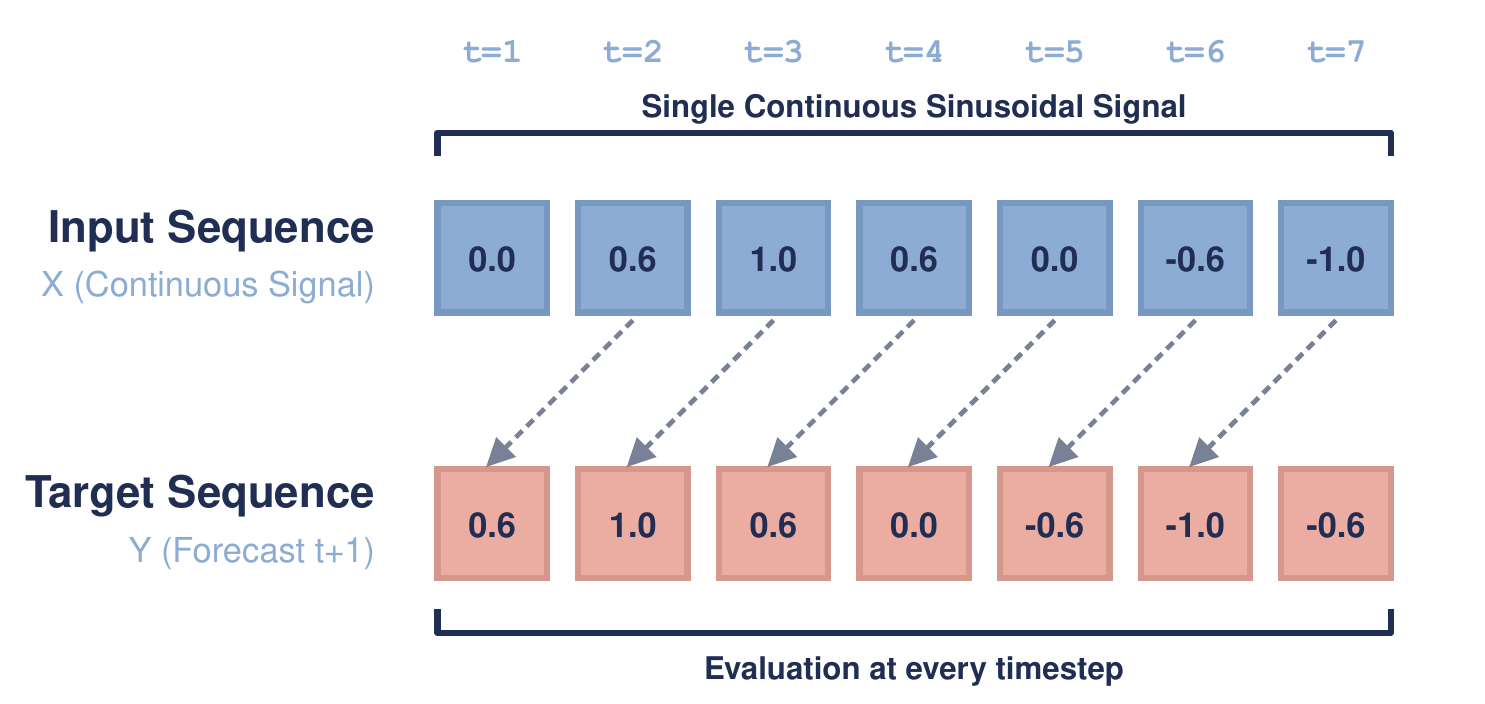}
        \caption{Illustration of Sinus Forecasting.}
        \label{fig:sinus}
    \end{minipage}
    \begin{minipage}[c]{0.48\textwidth}
        In this very simple task, the model is required to predict the future evolution of a sinusoidal signal. The difficulty is scaled by simultaneously extending the sequence length (e.g., from 200 to 2000) and the target forecast horizon (e.g., from 5 to 15).
    \end{minipage}
\end{figure}
\vspace{-15pt}

\subsubsection{Chaotic Forecasting}
\vspace{-25pt}
\noindent
\begin{figure}[htbp] 
    \centering
    \begin{minipage}[c]{0.48\textwidth}
        \centering
        \includegraphics[width=1.00\linewidth]{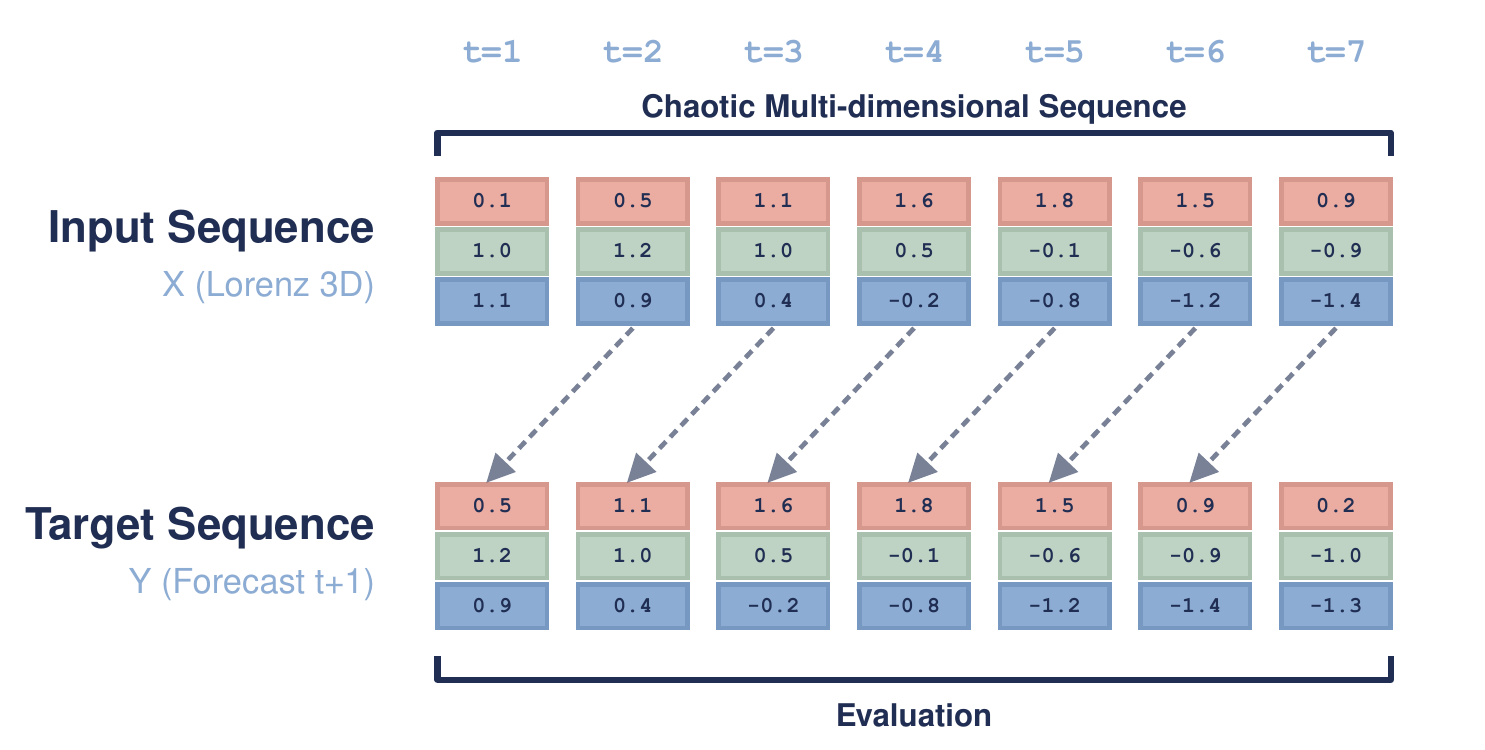}
        \caption{Illustration of Chaotic Forecasting.}
        \label{fig:chaotic}
    \end{minipage}
    \begin{minipage}[c]{0.48\textwidth}
        This task evaluates the model's ability to model chaotic dynamics by predicting the future state of a three-dimensional chaotic system based on the Lorenz equations \cite{lorenz2017deterministic}. Similar to the sinusoidal task, the difficulty can be scaled by increasing both the sequence length provided to the model and the forecast horizon.
    \end{minipage}
\end{figure}
\vspace{-15pt}

\subsection{Memory and Retention}

\subsubsection{Discrete and Continuous Postcasting}
\vspace{-25pt}
\noindent
\begin{figure}[htbp] 
    \centering
    \begin{minipage}[c]{0.48\textwidth}
        \centering
        \includegraphics[width=1.00\linewidth]{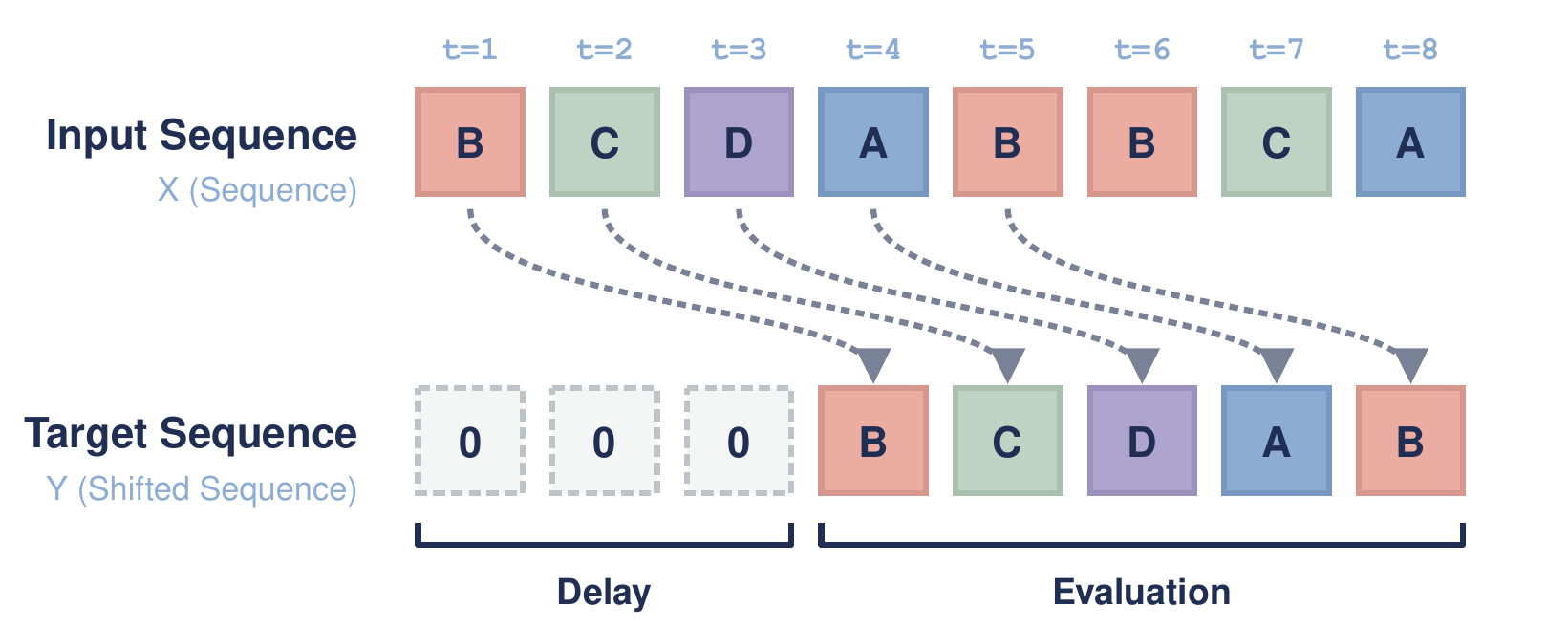}
        \caption{Illustration of Discrete Postcasting.}
        \label{fig:discrete_postcasting}
    \end{minipage}
    \begin{minipage}[c]{0.48\textwidth}
        \centering
        \includegraphics[width=1.00\linewidth]{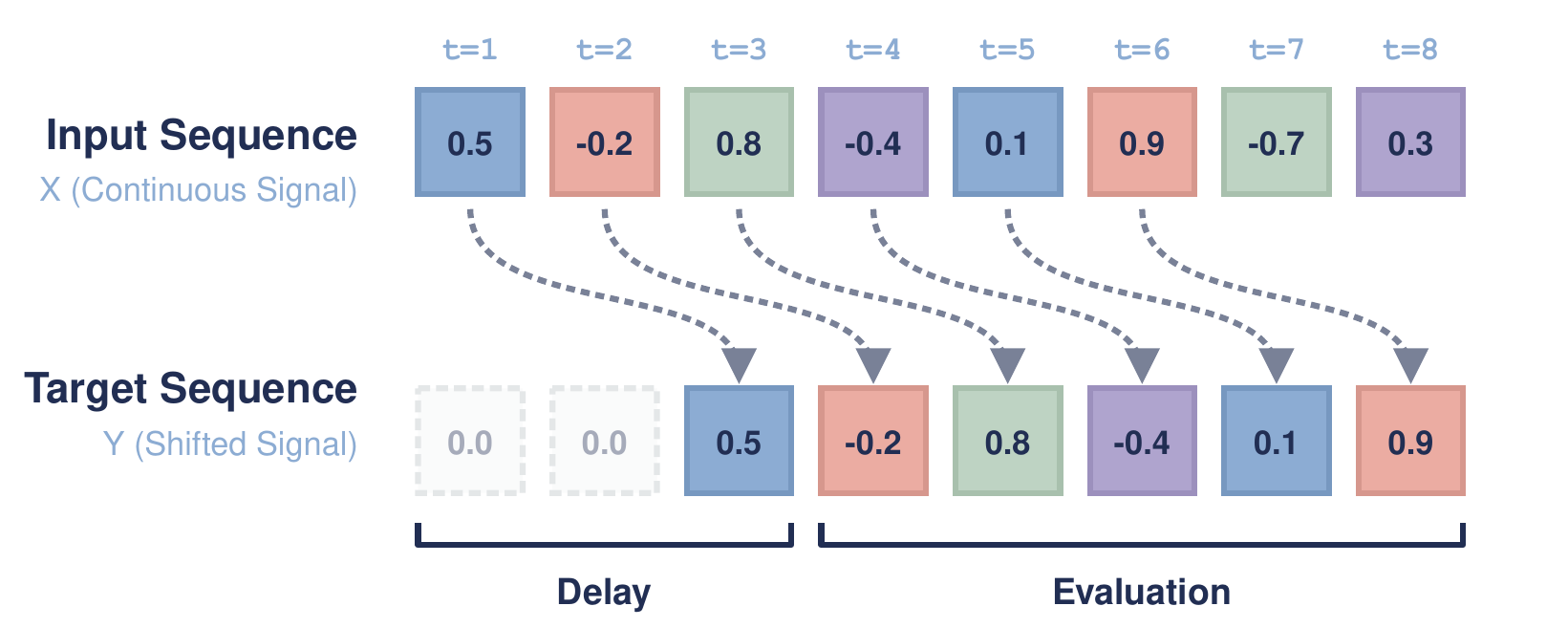}
        \caption{Illustration of Continuous Postcasting.}
        \label{fig:constinuous_postcasting}
    \end{minipage}
\end{figure}
\vspace{-15pt}

Inspired from the Memory Capacity defined by Jaeger \cite{jaeger2001echo}, postcasting is a pure temporal delay task. The model receives an input sequence and must reproduce it identically after a specified time shift. The scaling mechanism increases the sequence length and the retention delay (e.g., from 5 to 15 timesteps), while the discrete variant also scales the  vocabulary size (e.g., from 3 to 8 symbols).

\newpage

\subsubsection{Simple Copy}
\noindent
\begin{figure}[htbp] 
    \centering
    \begin{minipage}[c]{0.48\textwidth}
        \centering
        \includegraphics[width=1.00\linewidth]{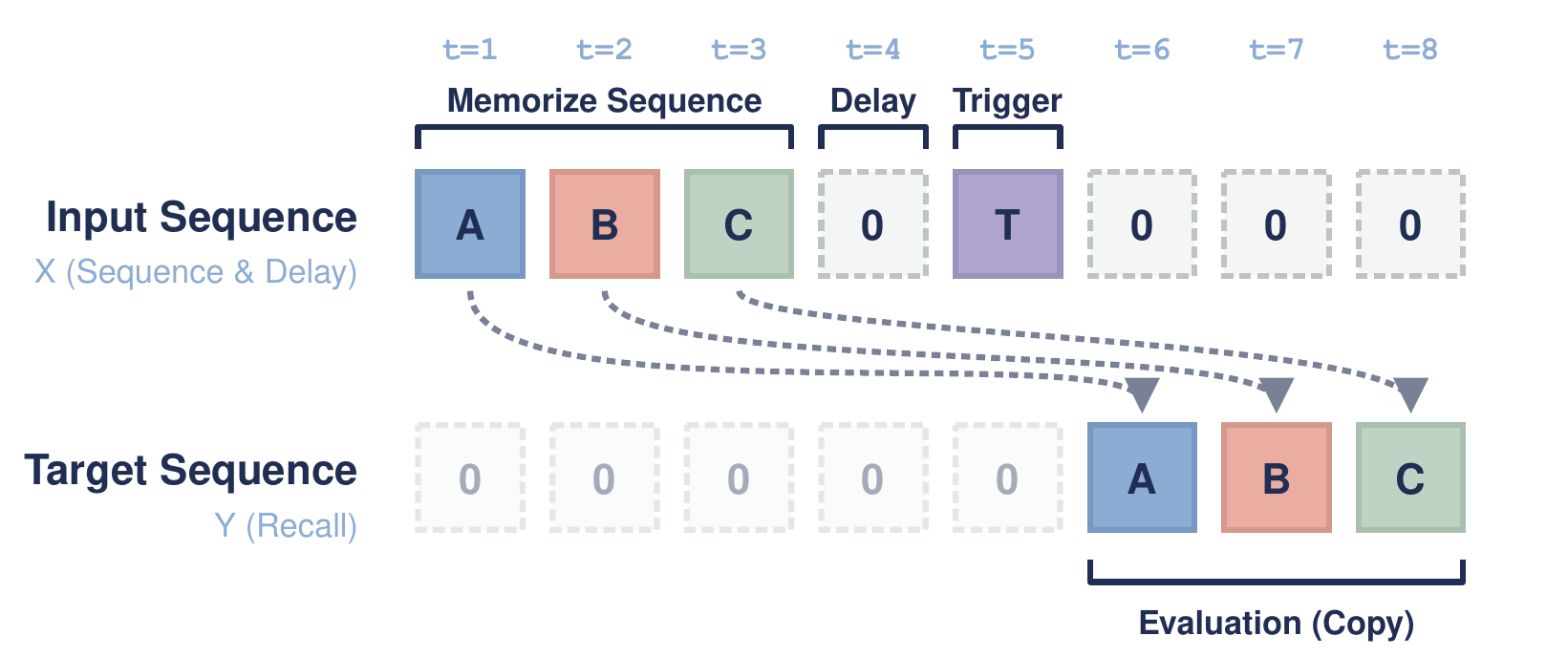}
        \caption{Illustration of Simple Copy.}
        \label{fig:simple_copy}
    \end{minipage}
    \begin{minipage}[c]{0.48\textwidth}
        Inspired from \cite{graves2014neural}, the simple copy task requires the model to read an entire sequence, hold it in memory during a silent delay period, and then reproduce the sequence in its entirety after a specific trigger token. Task difficulty is scaled up by increasing the sequence length, extending the waiting delay before the trigger appears, and expanding the vocabulary size.
    \end{minipage}
\end{figure}
\vspace{-15pt}

\subsubsection{Selective Copy}
\vspace{-25pt}
\noindent
\begin{figure}[htbp] 
    \centering
    \begin{minipage}[c]{0.48\textwidth}
        \centering
        \includegraphics[width=1.00\linewidth]{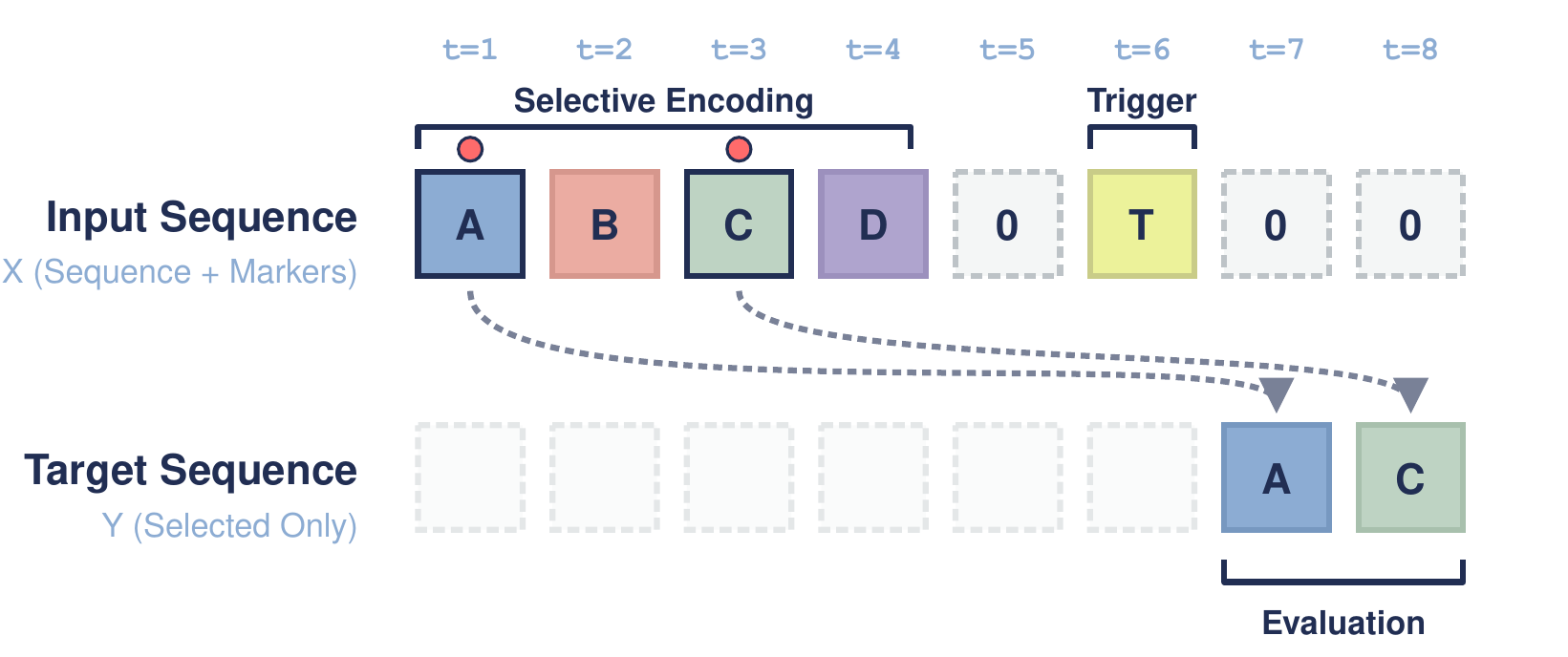}
        \caption{Illustration of Selective Copy.}
        \label{fig:selective_copy}
    \end{minipage}
    \begin{minipage}[c]{0.48\textwidth}
        Building upon the \texttt{simple copy} task and inspired from \cite{gu2023mamba}, this task introduces distraction elements. The model must memorize only a specific subset of marked tokens within a larger sequence and output only those targeted elements at the end. In addition to scaling the delay and vocabulary, the difficulty can be increased by scaling the number of target elements to retain (e.g., from 5 to 10 markers).
    \end{minipage}
\end{figure}
\vspace{-15pt}

\subsubsection{Associative Recall}
\vspace{-25pt}
\noindent
\begin{figure}[htbp] 
    \centering
    \begin{minipage}[c]{0.48\textwidth}
        \centering
        \includegraphics[width=1.00\linewidth]{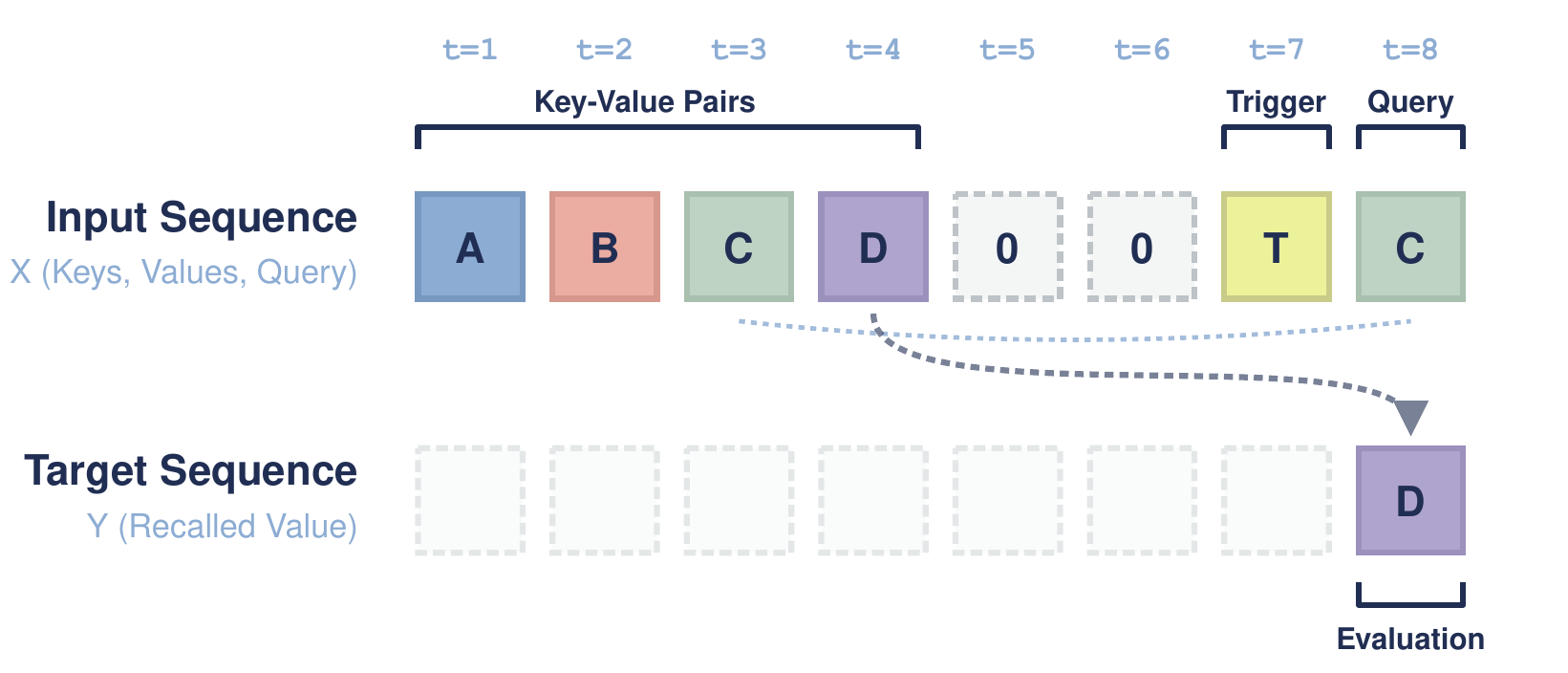}
        \caption{Illustration of Associative Recall.}
        \label{fig:associative_recall}
    \end{minipage}
    \begin{minipage}[c]{0.48\textwidth}
        Also inspired from \cite{graves2014neural}, this task tests associative memory by presenting the model with a sequence of key-value pairs. At the conclusion of the sequence, a seen key is provided as a query, and the model must retrieve and predict the corresponding associated value. The scaling mechanism adjusts the vocabulary size, the total number of pairs to memorize (e.g., from 3 to 8), and the overall sequence length.
    \end{minipage}
\end{figure}
\vspace{-15pt}

\subsection{Pattern Recognition and Completion}

\subsubsection{Discrete and Continuous Pattern Completion}
\noindent
\begin{figure}[H] 
    \centering
    \begin{minipage}[c]{0.48\textwidth}
        \centering
        \includegraphics[width=1.00\linewidth]{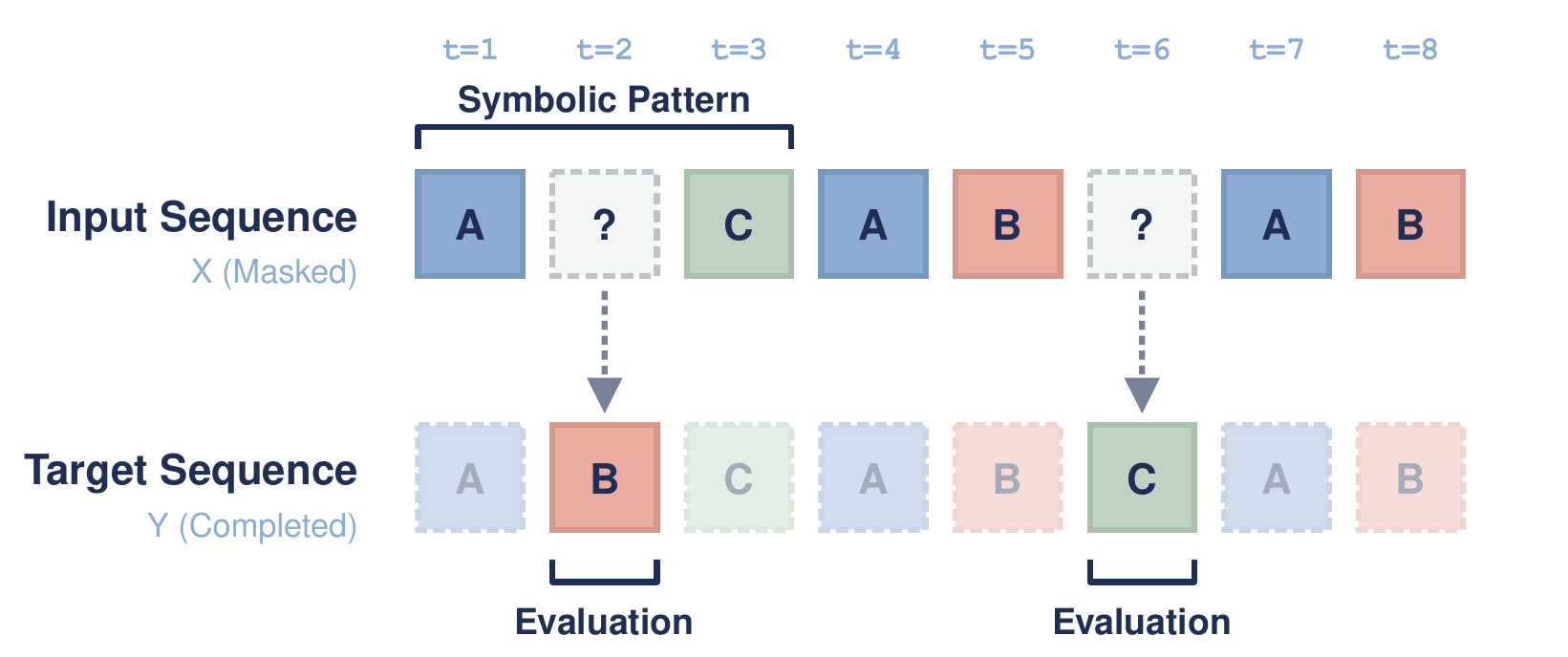}
        \caption{Illustration of D. Pattern Completion.}
        \label{fig:discrete_pattern_completion}
    \end{minipage}
    \begin{minipage}[c]{0.48\textwidth}
        \centering
        \includegraphics[width=1.00\linewidth]{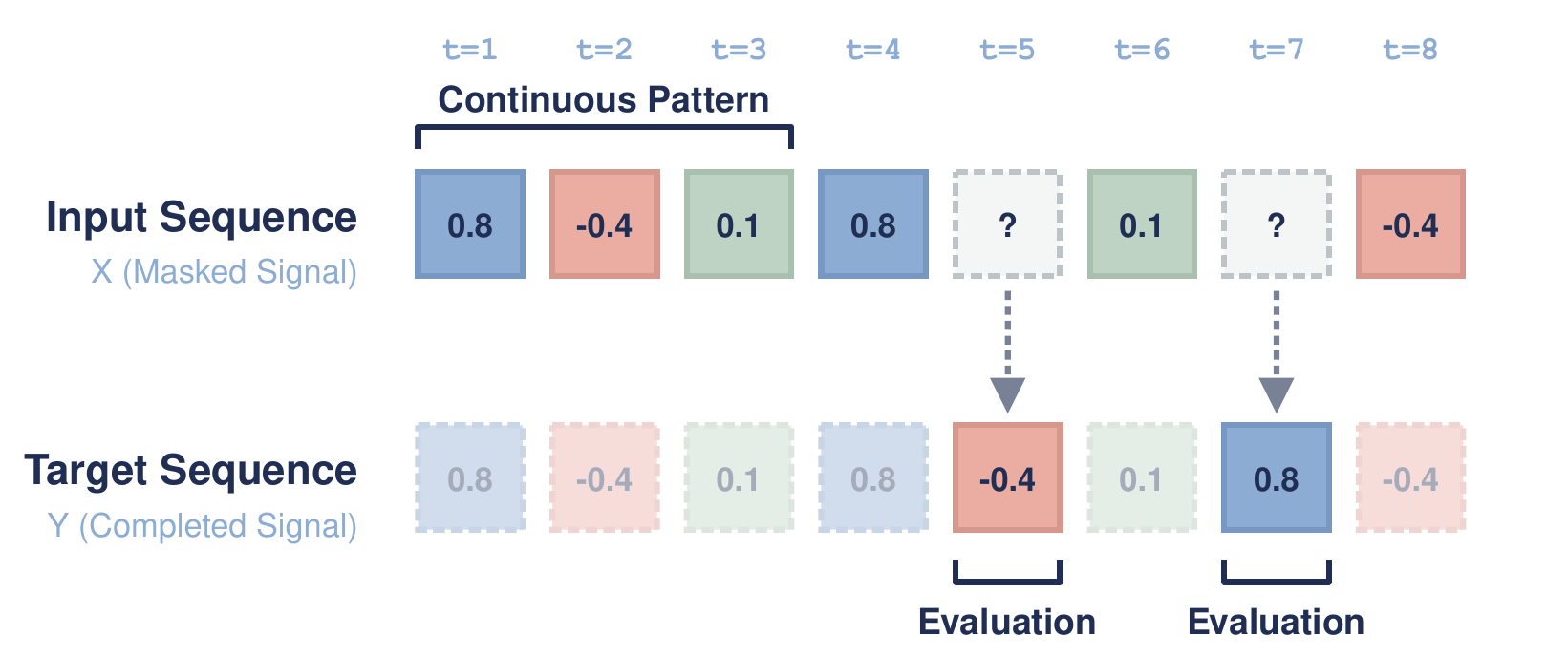}
        \caption{Illustration of C. Pattern Completion.}
        \label{fig:continuous_pattern_completion}
    \end{minipage}
\end{figure}

In these tasks, the model is exposed to a sequence with a periodic motif, but certain values within the pattern are masked. The objective is to identify the underlying pattern to successfully infer the missing components. The complexity scales by increasing the base length of the repeating motif (e.g., from 4 to 10), the total sequence length, and the vocabulary size for the discrete version.
\vspace{15pt}


\subsubsection{Induction Heads}
\noindent
\begin{figure}[htbp] 
    \centering
    \begin{minipage}[c]{0.48\textwidth}
        \centering
        \includegraphics[width=1.00\linewidth]{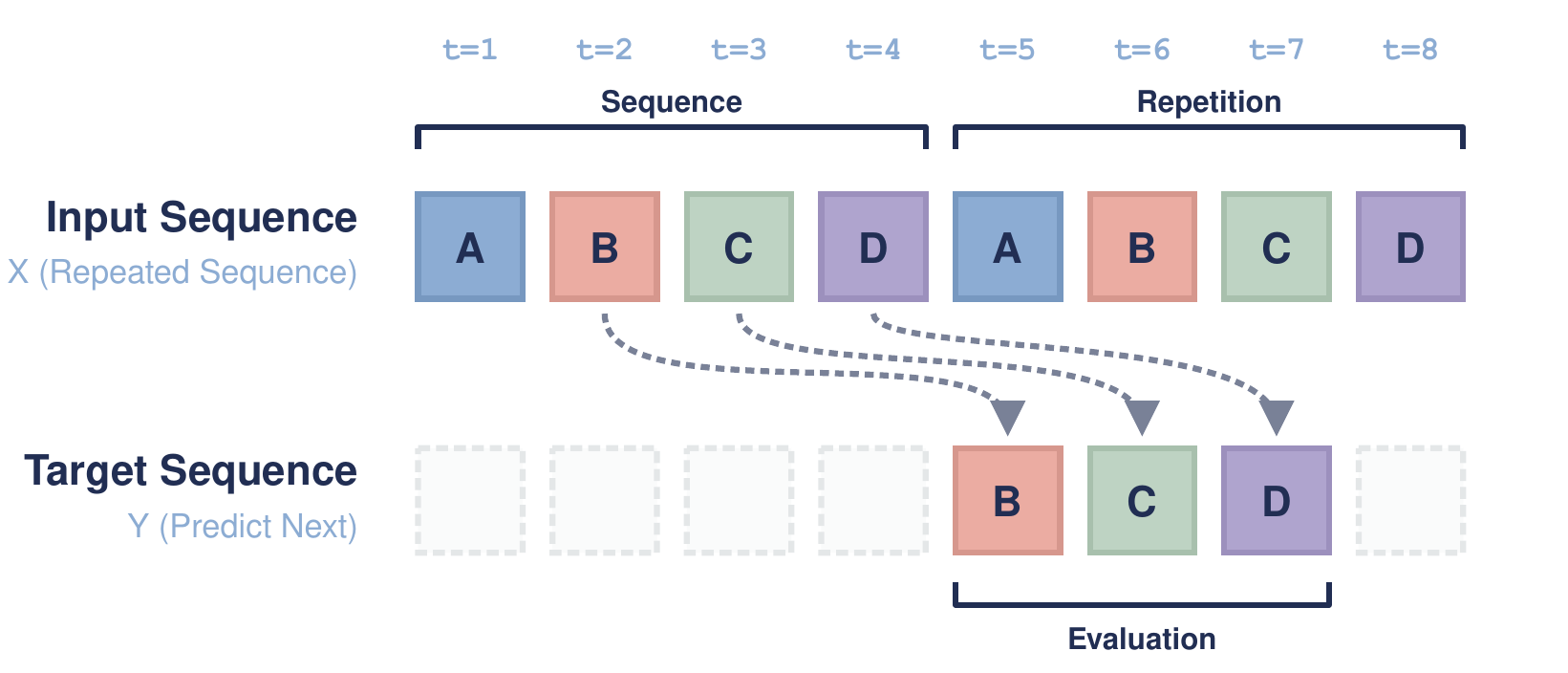}
        \caption{Illustration of Induction Heads.}
        \label{fig:induction_heads}
    \end{minipage}
    \begin{minipage}[c]{0.48\textwidth}
        Inspired from the in-context learning mechanisms observed in Transformer architectures \cite{olsson2022context}, this task presents a sequence where the second half is an exact duplicate of the first half. The model must recognize this structure to predict the next token in the copied sequence. The difficulty can be increased by expanding the sequence length and increasing the vocabulary size.
    \end{minipage}
\end{figure}
\vspace{-15pt}

\subsection{Reasoning and Algorithmic Manipulation}
\subsubsection{Adding Problem}
\vspace{-10pt}
\noindent
\begin{figure}[htbp] 
    \centering
    \begin{minipage}[c]{0.48\textwidth}
        \centering
        \includegraphics[width=1.00\linewidth]{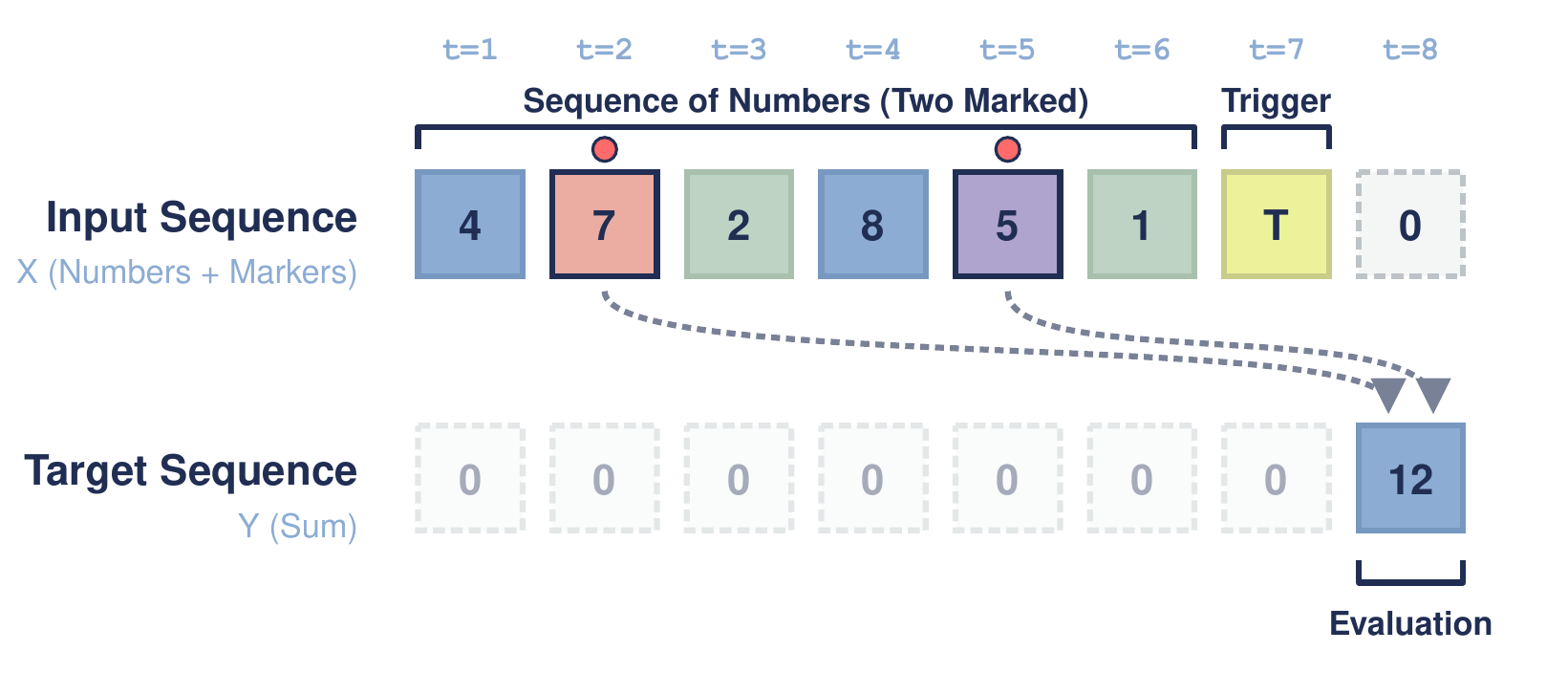}
        \caption{Illustration of Adding Problem.}
        \label{fig:adding_problem}
    \end{minipage}
    \begin{minipage}[c]{0.48\textwidth}
        Inspired from \cite{arjovsky2016unitary}, the model receives a sequence of random numbers as symbols along with positional markers. Once a trigger signal is received, it must compute and output the sum of only the marked numbers. Scaling increases the sequence length, making the localization of markers more challenging, and raises the maximum possible value of the numbers to be added (e.g., from 3 to 8), effectively increasing the vocabulary size.
    \end{minipage}
\end{figure}

\subsubsection{Sorting Problem}
\vspace{-10pt}
\noindent
\begin{figure}[htbp] 
    \centering
    \begin{minipage}[c]{0.48\textwidth}
        \centering
        \includegraphics[width=1.00\linewidth]{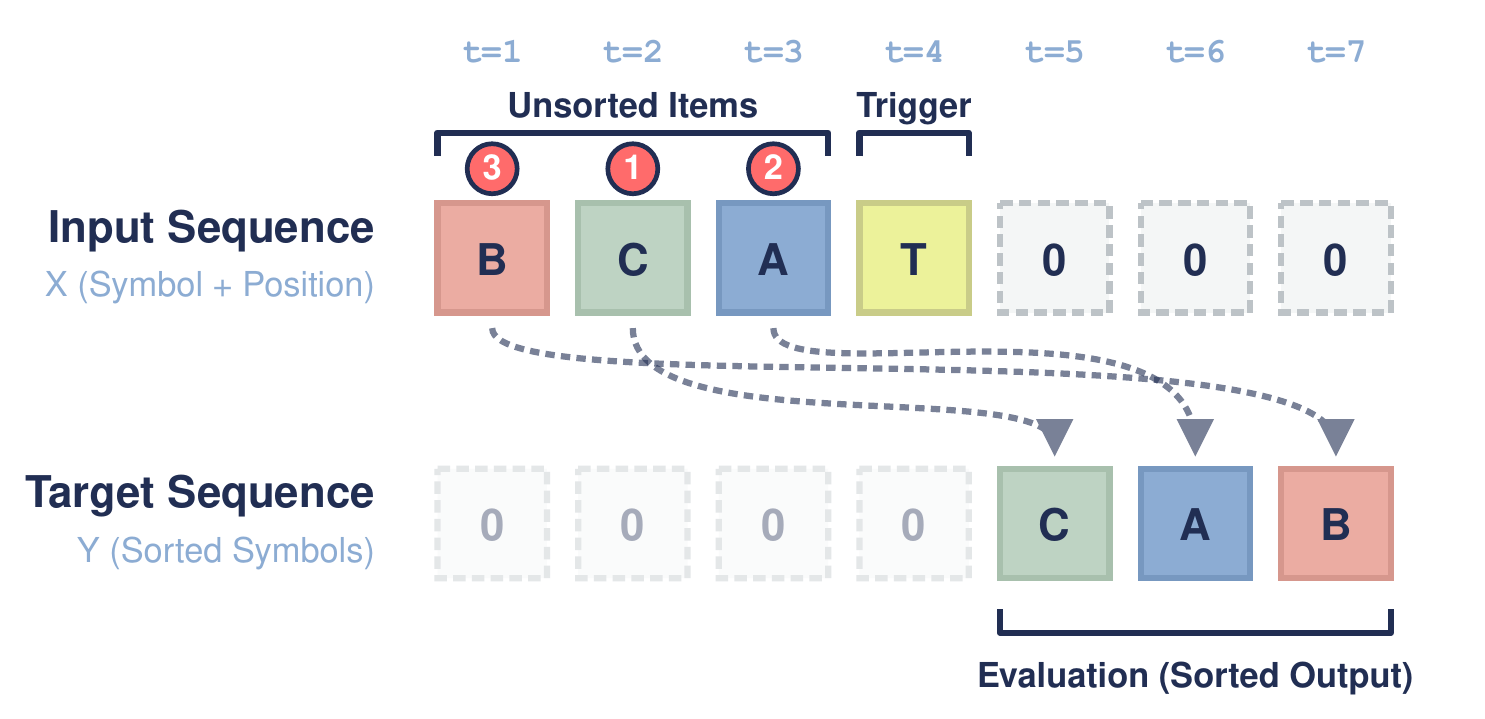}
        \caption{Illustration of Sorting Problem.}
        \label{fig:sorting_problem}
    \end{minipage}
    \begin{minipage}[c]{0.48\textwidth}
        The model is provided with a random sequence of symbols, each paired with a randomized target position. After a trigger signal, the model must output the entire sequence sorted into the correct positional order. Scaling this task involves increasing both the sequence length and the vocabulary size, which exponentially inflates the number of possible positional permutations.
    \end{minipage}
\end{figure}

\subsubsection{Bracket Matching}
\vspace{-10pt}
\noindent
\begin{figure}[H] 
    \centering
    \begin{minipage}[c]{0.48\textwidth}
        \centering
        \includegraphics[width=1.00\linewidth]{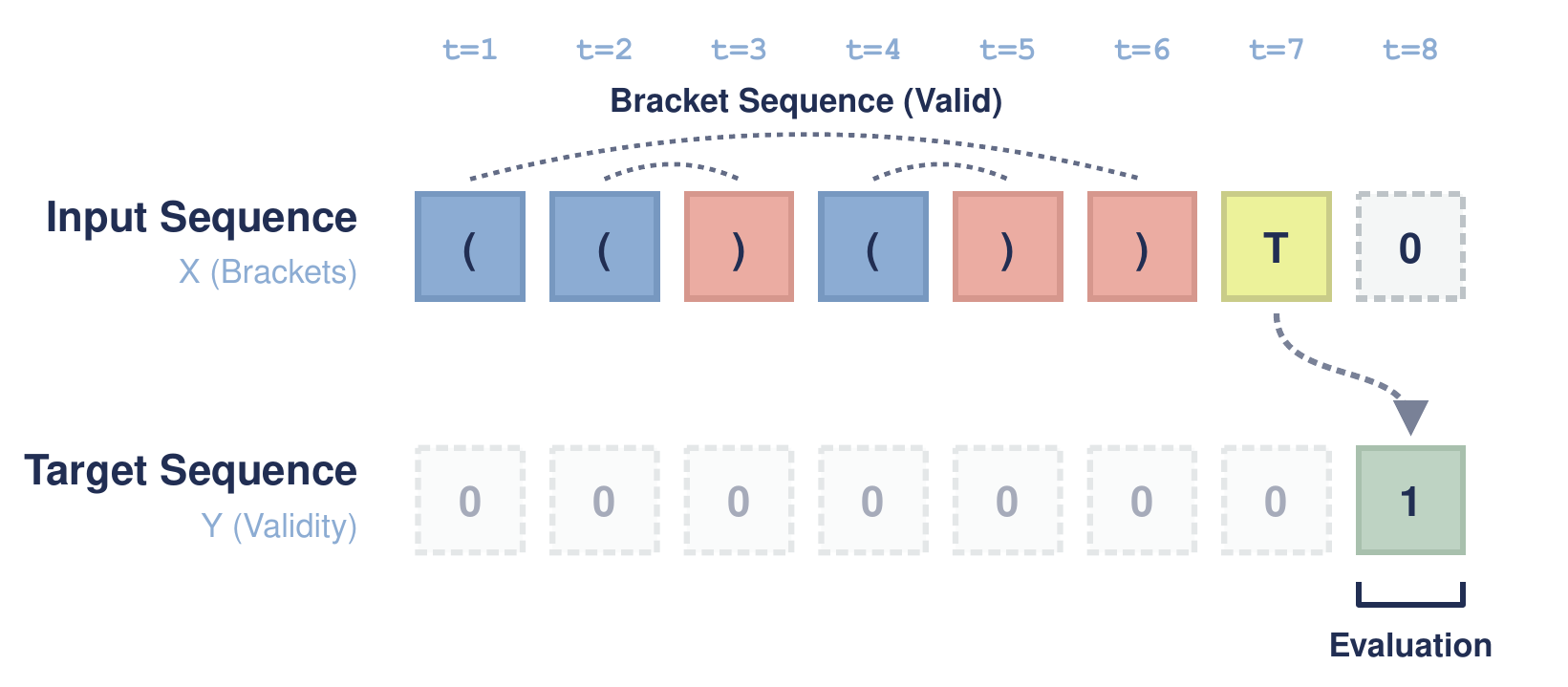}
        \caption{Illustration of Bracket Matching.}
        \label{fig:bracket_matching}
    \end{minipage}
    \begin{minipage}[c]{0.48\textwidth}
        This task evaluates hierarchical context maintenance. The sequence consists of opening and closing parentheses with random mutations, and the model must determine at the final timestep whether the entire string represents a valid hierarchy. The scaling mechanism increases both the total sequence length and the maximum allowable depth of the bracket hierarchy (e.g., from 5 to 10).
    \end{minipage}
\end{figure}

\subsubsection{Cross Situation}
\noindent
\begin{figure}[htbp] 
    \centering
    \begin{minipage}[c]{0.48\textwidth}
        \centering
        \includegraphics[width=1.00\linewidth]{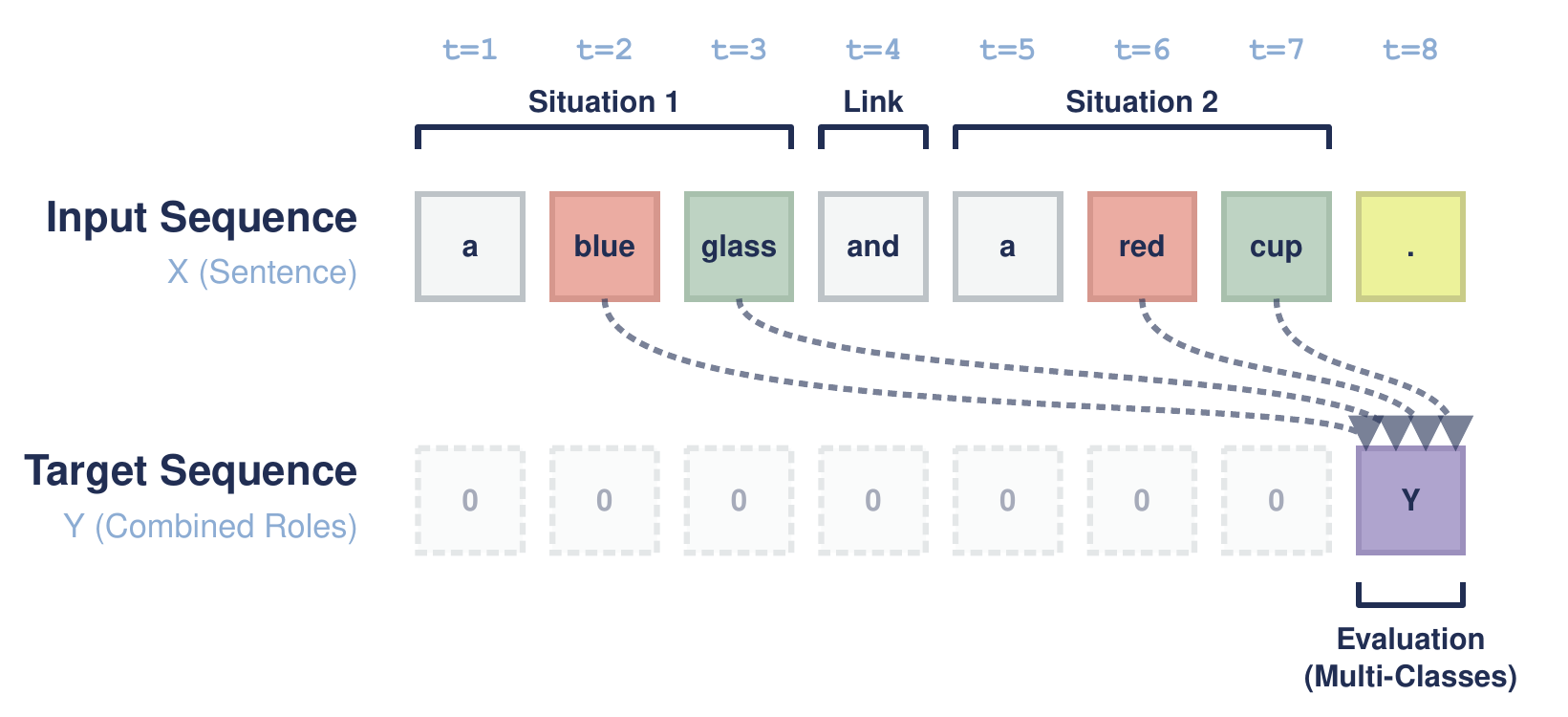}
        \caption{Illustration of Cross Situation.}
        \label{fig:cross_situation}
    \end{minipage}
    \begin{minipage}[c]{0.48\textwidth}
        Defined in \cite{juven2020cross, variengien2020journey} as a simplified natural language reasoning problem encoded in one-hot vectors, Cross Situation requires the model to read a sentence describing two crossed situations involving objects, colors, and positions. It must then infer the correct roles and attributes at the end of the sequence. The task features linguistic complexities such as synonyms (e.g., "center" and "middle" representing the same spatial label) and polysemous words (e.g., "orange" representing both a color and a fruit object). The difficulty scales by expanding the entities available, shifting from a restricted vocabulary (e.g., 2 objects, 2 colors, 2 positions) to a broader one (e.g., 8 objects, 8 colors, 8 positions). 
        
    \end{minipage}
\end{figure}
\vspace{-15pt}

\section{Experimental Setup}

To guarantee a fair comparison across fundamentally diverse architectures, we employ a strict parameter matching protocol. Rather than relying on arbitrary default hyperparameter configurations that might inadvertently favor one model over another, we utilize a binary search algorithm to dynamically adjust the hidden dimensions. This algorithm forces every architecture to adhere to predefined parameter budgets of 1k, 10k, and 100k trainable parameters. Furthermore, to ensure statistical robustness and mitigate the high variance often introduced by random weight initialization in small models, every experimental configuration is evaluated across 10 seeds, on two pre-configured difficulty levels: \texttt{small (SM)} and \texttt{medium (MD)} (refer to Appendix \ref{app:task_config} for full tasks configurations).
For all differentiable models (GRU, LSTM, xLSTM, Transformer Decoder-Only (DO) and Transformer Encoder-Decoder (ED)), training is conducted using PyTorch \cite{paszke2019pytorch} with the Adam optimizer \cite{kingma2014adam}, utilizing a fixed batch size of 10 samples, a maximum number of 200 epochs and an early stopping of 10 epochs to prevent overfitting and ensure fair comparison across all architectures. We also manage sequence packing and padding, which is particularly crucial for architectures like xLSTM that typically operate on unpadded streams. Since deep learning models are sensitive to optimization hyperparameters at smaller scales, we perform a grid search over five distinct learning rates ($10^{-2}, 3 \times 10^{-3}, 10^{-3}, 3 \times 10^{-4}, 10^{-4}$). Each learning rate is tested across the 10 seeds, resulting in 50 independent training runs per task, difficulty level, and parameter size combination. 
A model's peak performance is highlighted through the best overall score, representing the single best result achieved across all 150 training runs (10 seeds x 5 learning rates x 3 model sizes). This metric is used in our Cognitive Radar (Figure \ref{fig:radar_charts}). Furthermore, we report the mean and standard deviation of the test scores for the optimal configuration, determined by selecting the learning rate that yields the best average validation score across its 10 seeds. Full score descriptions for both measures are available in Appendix \ref{app:detailed_results}.
The Echo State Network (ESN) baseline provides a radically different computational paradigm, as it uses a fixed, randomized dynamical reservoir and does not rely on backpropagation. Consequently, standard deep learning training protocols are inapplicable. Instead, we implement a large, parallelized hyperparameter search protocol to optimize the reservoir's dynamics. This search explores thousands of configurations across three critical hyperparameters: the leaking rate, the spectral radius, and the input scaling. For each reservoir configuration, the optimal Ridge for the linear readout layer is automatically selected on the validation set. All the source code required to reproduce all the experiments is publicly available.\footnote{\url{https://anonymous.4open.science/r/CogScale/}} 
\newpage
\section{Results}

\subsection{Peak Cognitive Capabilities (The Cognitive Radar)}

\begin{figure}[ht]
    \centering
    \includegraphics[width=1\linewidth]{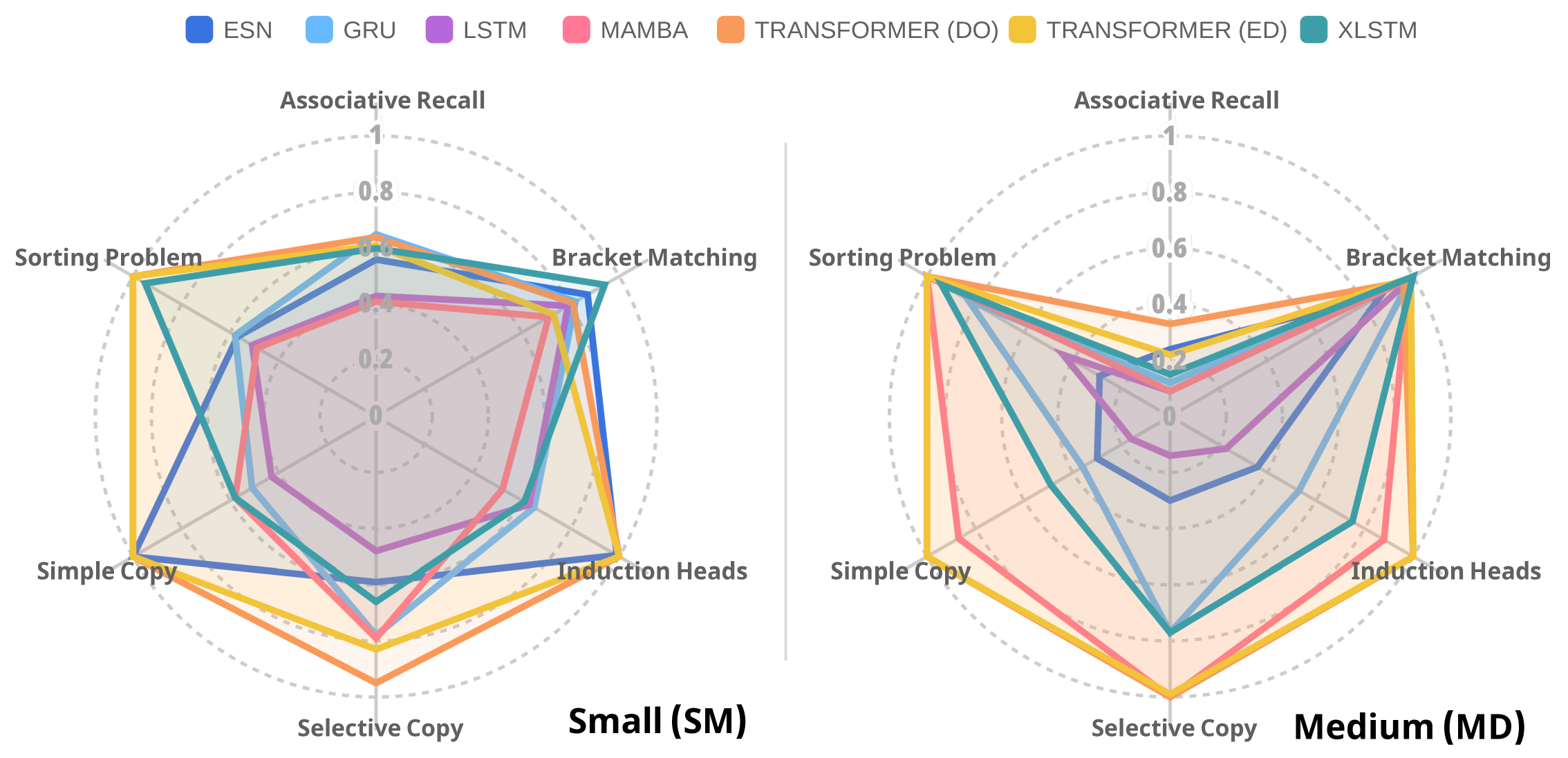}
    \caption{Cognitive Radar visualizing peak performance (Accuracy) for the seven baseline architectures across six selected tasks. Plots contrast architectural capabilities on \texttt{small} difficulty (left) and \texttt{medium} difficulty (right), based on the "best overall" scores (refer to Appendix \ref{app:detailed_results} for full score description). Larger colored areas signify superior cognitive abilities, illustrating how Transformers and modern State Space Models maintain better performance under increased difficulty compared to classical recurrent and reservoir computing models.}
    \label{fig:radar_charts}
\end{figure}


To evaluate the strengths of each architecture, we first analyze their peak cognitive capabilities using the "best overall" metric on the \texttt{small (SM)} and \texttt{medium (MD)} difficulty levels. By aggregating these optimal performances we construct a "Cognitive Radar" (Figure \ref{fig:radar_charts}) that provides an immediate visual profile of each model. 
We can notably observe that attention architectures (Transformer models) demonstrate a superiority in reasoning and manipulation, achieving perfect scores (0.00 error) on tasks such as \texttt{simple copy}, \texttt{adding problem}, and \texttt{induction heads}. 
The modern State Space Model, Mamba, is highly competitive with Transformers (DO and EC) across these six discriminative tasks, yet it generally remains slightly behind. This difference in performance is more noticeable for \texttt{medium} tasks. For instance, while Transformers maintain perfect or near-perfect accuracy on \texttt{simple copy} and \texttt{induction heads}, Mamba experiences a noticeable degradation in performance. In contrast, the Echo State Network (ESN) shows a distinct duality. While the ESN performs remarkably well on the \texttt{small} difficulty level tasks that require memory retention, like on \texttt{simple copy} or \texttt{induction heads}, it struggles when confronted with tasks requiring reasoning, selection, and manipulation, such as the \texttt{sorting problem} and \texttt{selective copy}.

\subsection{Impact of Task Difficulty (Small vs. Medium)}

Transitioning from the \texttt{small} to the \texttt{medium} difficulty level involves extending sequence lengths, increasing delays, expanding vocabulary sizes and increasing the number of samples which is ideal to tests the scalability of architectures. This increase in complexity causes a performance drop for several baseline models. For example, on \texttt{simple copy}, the ESN's error spikes drastically from 0.00  to 0.70, while the standard LSTM degrades from 0.57 to 0.84. A similar collapse is observed for the ESN on the \texttt{adding problem}, where its error jumps from 0.00 to 0.69. In contrast, modern architectures resist this difficulty scaling remarkably well. Transformers (both variants) and the xLSTM maintain near perfect error rates on the medium difficulty of the \texttt{sorting problem} and \texttt{bracket matching} tasks, demonstrating that their memory mechanisms are far more robust to increased difficulty than classical recurrent networks. On the other hand, Mamba really benefits from more samples, even if task difficulty is increased. 

\subsection{Scaling Behavior and Parameter Efficiency (1k to 100k)}

Since we tested different model sizes (1k, 10k, and 100k parameters), we can observe the scaling behavior of each architecture on 10 training seeds. Generally, scaling from 1k to 10k parameter brings significant and consistent performance gains across the majority of deep learning models. However, scaling to 100k parameters does not always guarantee better results on these synthetic tasks. We observe that standard and advanced recurrent architectures (LSTM, GRU, and xLSTM) often have an optimal parameter threshold beyond which performance degrades. For example, the GRU's performance on the \texttt{small adding problem} task falls from a 0.36 average error at 10k parameters to 0.55 at 100k parameters. On the other hand, attention architectures (Transformers) and State Space Models (Mamba) appear more resilient, stabilizing or continuing to improve their performance at the 100k scale without suffering from parameter inefficiency. Finally, it is important to note that the Echo State Network (ESN) was only evaluated for 1k and 10k parameter. Even so, we observe incredible error reductions when expanding the model size from 1k to 10k: the error rate drops from 0.47 to 0.02 on the \texttt{small adding problem}, from 0.57 to 0.10 on the medium \texttt{discrete pattern completion}, and from 0.63 to a perfect 0.00 on the medium \texttt{discrete postcasting}. Unfortunately, while backpropagation models naturally benefit from massive scaling, optimizing a 100k parameter ESN remains computationally expensive due to the hyperparameter search it requires, which prevents it from scaling efficiently.

\section{Discussion}

CogScale is designed to serve as a "sanity check" for new architecture elaboration. Our results show that the basic cognitive abilities, like retrieval, induction, and manipulation, are non-trivial prerequisites for any architecture aspiring to the status of a foundation model. Before spending thousands of GPU hours to train on massive datasets like The Pile or OpenWebText, an architecture must first prove its cognitive abilities at a small scale. We argue that a foundation model must, by definition, be "generalist". If an architecture, such as LSTM or ESN, fails to solve specific cognitive tasks like the \texttt{sorting problem} or \texttt{selective copy}, or fails to scale its performance across difficulty levels, it is highly unlikely that it will demonstrate emergent reasoning or robust generalization when scaled to millions, billions or even trillions of parameters. Among our evaluated baselines, only Transformers, Mamba, and to a significant extent xLSTM, emerge as legitimate candidates for massive NLP training, as they are the only models that maintain good performance across the full spectrum of CogScale tasks, at different scales.
On the other hand, our results show very interesting performance for the ESN at a small scale. While the ESN fails to perform across complex tasks, its randomly initialized and non-trained reservoir actually outperforms deep learning models mostly on forecasting, continuous and memory task. However, as soon as a task requires reasoning, manipulation, or precise information selection, its performance is less competitive. This shows that the absence of backpropagation prevents the model from developing the complex internal representations necessary for complex symbolic operations, ultimately confining it to the role of a specialist rather than a generalist.

The ability to identify such architectural limitations early in the development cycle is particularly important given the current race for better architectures. The reliance on large-scale training \cite{kaplan2020scaling, hoffmann2022training} has led to a research paradigm that is both ecologically damaging \cite{bender2021dangers, varoquaux2025hype} and economically expensive for the majority of academic laboratories. CogScale proposes a more sustainable alternative by demonstrating that architectures can be effectively discriminated at a much smaller scale, specifically between 1k and 100k parameters. By providing a test protocol for new architectures, CogScale prevents the training of deficient models on massive supercomputers. 
Filtering out these architectures offers a substantial opportunity to reduce energy consumption and computational costs, while making architectural research more accessible to smaller teams that do not have the extensive resources of industrial AI labs.
Beyond research and development, this focus on lower parameter budgets has immediate practical implications for real-world problems where massive foundation models are not suitable~\cite{varoquaux2025hype}. In industrial edge computing and healthcare devices, models are often constrained by memory and power limitations. Not every application requires a trillion parameter LLM; many require only the ability to perform specific reasoning and memory task with minimal overhead \cite{chen2024role}. Our scaling analysis reveals that architectures like Transformers and Mamba achieve excellent performance even at the 10k parameter scale, and that simpler models like ESNs can outperform them at 1k parameter scale. Consequently, CogScale provides a clear roadmap for engineers in these sectors, allowing them to select the most efficient architecture based on strict hardware constraints.

\section{Conclusion}

CogScale offers a lightweight alternative to the current race toward massive models, which has made the evaluation of new architectures increasingly slow and costly.
This framework of 14 tasks serves as a "sanity check" to validate the basic cognitive abilities required for generalization. Our evaluation demonstrates that while models like the ESN provide impressive performance at the 1k parameter scale, only Transformers, Mamba, and xLSTM prove to be ‘‘generlist‘‘ architectures, scaling effectively to complex compositional tasks that combine various rules and operations.. 
Although this synthetic benchmark should be complemented by evaluations on large-scale datasets such as OWT or The Pile, it successfully filters out deficient models at an early stage, helping researchers save valuable time and financial resources while avoiding unnecessary energy consumption.


\begin{ack}
Experiments presented in this paper were performed using the \textit{PlaFRIM} HPC cluster, supported by Inria Bordeaux.

We would also like to thank \textit
{Romain de Coudenhove} for his key contribution during his internship. He developed the highly optimized framework that enabled the automated hyperparameter (HP) search for Reservoir Computing across all tasks, scales, and difficulty levels presented in this work.
\end{ack}

\newpage
\bibliographystyle{unsrt} 
\bibliography{ref} 

\begin{thebibliography}{10}

\bibitem{paperno2016lambada}
Denis Paperno, Germ{\'a}n Kruszewski, Angeliki Lazaridou, Ngoc-Quan Pham,
  Raffaella Bernardi, Sandro Pezzelle, Marco Baroni, Gemma Boleda, and Raquel
  Fern{\'a}ndez.
\newblock The lambada dataset: Word prediction requiring a broad discourse
  context.
\newblock In {\em Proceedings of the 54th annual meeting of the association for
  computational linguistics (volume 1: Long papers)}, pages 1525--1534, 2016.

\bibitem{sakaguchi2021winogrande}
Keisuke Sakaguchi, Ronan~Le Bras, Chandra Bhagavatula, and Yejin Choi.
\newblock Winogrande: An adversarial winograd schema challenge at scale.
\newblock {\em Communications of the ACM}, 64(9):99--106, 2021.

\bibitem{bisk2020piqa}
Yonatan Bisk, Rowan Zellers, Jianfeng Gao, Yejin Choi, et~al.
\newblock Piqa: Reasoning about physical commonsense in natural language.
\newblock In {\em Proceedings of the AAAI conference on artificial
  intelligence}, volume~34, pages 7432--7439, 2020.

\bibitem{zellers2019hellaswag}
Rowan Zellers, Ari Holtzman, Yonatan Bisk, Ali Farhadi, and Yejin Choi.
\newblock Hellaswag: Can a machine really finish your sentence?
\newblock In {\em Proceedings of the 57th annual meeting of the association for
  computational linguistics}, pages 4791--4800, 2019.

\bibitem{Gokaslan2019OpenWeb}
Aaron Gokaslan and Vanya Cohen.
\newblock Openwebtext corpus.
\newblock \url{http://Skylion007.github.io/OpenWebTextCorpus}, 2019.

\bibitem{gao2020pile}
Leo Gao, Stella Biderman, Sid Black, Laurence Golding, Travis Hoppe, Charles
  Foster, Jason Phang, Horace He, Anish Thite, Noa Nabeshima, et~al.
\newblock The pile: An 800gb dataset of diverse text for language modeling.
\newblock {\em arXiv preprint arXiv:2101.00027}, 2020.

\bibitem{strubell2019energy}
Emma Strubell, Ananya Ganesh, and Andrew McCallum.
\newblock Energy and policy considerations for deep learning in nlp.
\newblock In {\em Proceedings of the 57th annual meeting of the association for
  computational linguistics}, pages 3645--3650, 2019.

\bibitem{patterson2021carbon}
David Patterson, Joseph Gonzalez, Quoc Le, Chen Liang, Lluis-Miquel Munguia,
  Daniel Rothchild, David So, Maud Texier, and Jeff Dean.
\newblock Carbon emissions and large neural network training.
\newblock {\em arXiv preprint arXiv:2104.10350}, 2021.

\bibitem{schwartz2020green}
Roy Schwartz, Jesse Dodge, Noah~A Smith, and Oren Etzioni.
\newblock Green ai.
\newblock {\em Communications of the ACM}, 63(12):54--63, 2020.

\bibitem{varoquaux2025hype}
Ga{\"e}l Varoquaux, Sasha Luccioni, and Meredith Whittaker.
\newblock Hype, sustainability, and the price of the bigger-is-better paradigm
  in {AI}.
\newblock In {\em Proceedings of the 2025 ACM Conference on Fairness,
  Accountability, and Transparency}, pages 61--75, 2025.

\bibitem{wang2024tssurvey}
Yuxuan Wang, Haixu Wu, Jiaxiang Dong, Yong Liu, Mingsheng Long, and Jianmin
  Wang.
\newblock Deep time series models: A comprehensive survey and benchmark.
\newblock 2024.

\bibitem{hochreiter1997long}
Sepp Hochreiter and J{\"u}rgen Schmidhuber.
\newblock Long short-term memory.
\newblock {\em Neural computation}, 9(8):1735--1780, 1997.

\bibitem{gers2000learning}
Felix~A Gers, J{\"u}rgen Schmidhuber, and Fred Cummins.
\newblock Learning to forget: Continual prediction with lstm.
\newblock {\em Neural computation}, 12(10):2451--2471, 2000.

\bibitem{cho2014learning}
Kyunghyun Cho, Bart Van~Merri{\"e}nboer, {\c{C}}a{\u{g}}lar Gul{\c{c}}ehre,
  Dzmitry Bahdanau, Fethi Bougares, Holger Schwenk, and Yoshua Bengio.
\newblock Learning phrase representations using rnn encoder--decoder for
  statistical machine translation.
\newblock In {\em Proceedings of the 2014 conference on empirical methods in
  natural language processing (EMNLP)}, pages 1724--1734, 2014.

\bibitem{vaswani2017attention}
Ashish Vaswani, Noam Shazeer, Niki Parmar, Jakob Uszkoreit, Llion Jones,
  Aidan~N Gomez, {\L}ukasz Kaiser, and Illia Polosukhin.
\newblock Attention is all you need.
\newblock {\em Advances in neural information processing systems}, 30, 2017.

\bibitem{radford2018improving}
Alec Radford, Karthik Narasimhan, Tim Salimans, Ilya Sutskever, et~al.
\newblock Improving language understanding by generative pre-training.
\newblock 2018.

\bibitem{gu2023mamba}
Albert Gu and Tri Dao.
\newblock Mamba: Linear-time sequence modeling with selective state spaces.
\newblock {\em arXiv preprint arXiv:2312.00752}, 2023.

\bibitem{beck2024xlstm}
Maximilian Beck, Korbinian P{\"o}ppel, Markus Spanring, Andreas Auer,
  Oleksandra Prudnikova, Michael Kopp, G{\"u}nter Klambauer, Johannes
  Brandstetter, and Sepp Hochreiter.
\newblock xlstm: Extended long short-term memory.
\newblock {\em Advances in Neural Information Processing Systems},
  37:107547--107603, 2024.

\bibitem{jaeger2001echo}
Herbert Jaeger.
\newblock The “echo state” approach to analysing and training recurrent
  neural networks-with an erratum note.
\newblock {\em Bonn, Germany: German national research center for information
  technology gmd technical report}, 148(34):13, 2001.

\bibitem{jaeger2002adaptive}
Herbert Jaeger.
\newblock Adaptive nonlinear system identification with echo state networks.
\newblock {\em Advances in neural information processing systems}, 15, 2002.

\bibitem{tay2020long}
Yi~Tay, Mostafa Dehghani, Samira Abnar, Yikang Shen, Dara Bahri, Philip Pham,
  Jinfeng Rao, Liu Yang, Sebastian Ruder, and Donald Metzler.
\newblock Long range arena: A benchmark for efficient transformers.
\newblock {\em arXiv preprint arXiv:2011.04006}, 2020.

\bibitem{weston2015towards}
Jason Weston, Antoine Bordes, Sumit Chopra, Alexander~M Rush, Bart
  Van~Merri{\"e}nboer, Armand Joulin, and Tomas Mikolov.
\newblock Towards ai-complete question answering: A set of prerequisite toy
  tasks.
\newblock {\em arXiv preprint arXiv:1502.05698}, 2015.

\bibitem{lukovsevivcius2009reservoir}
Mantas Luko{\v{s}}evi{\v{c}}ius and Herbert Jaeger.
\newblock Reservoir computing approaches to recurrent neural network training.
\newblock {\em Computer science review}, 3(3):127--149, 2009.

\bibitem{yan2024emerging}
Min Yan, Can Huang, Peter Bienstman, Peter Tino, Wei Lin, and Jie Sun.
\newblock Emerging opportunities and challenges for the future of reservoir
  computing.
\newblock {\em Nature Communications}, 15(1):2056, 2024.

\bibitem{lorenz2017deterministic}
Edward~N Lorenz.
\newblock Deterministic nonperiodic flow 1.
\newblock In {\em Universality in Chaos, 2nd edition}, pages 367--378.
  Routledge, 2017.

\bibitem{graves2014neural}
Alex Graves, Greg Wayne, and Ivo Danihelka.
\newblock Neural turing machines.
\newblock {\em arXiv preprint arXiv:1410.5401}, 2014.

\bibitem{olsson2022context}
Catherine Olsson, Nelson Elhage, Neel Nanda, Nicholas Joseph, Nova DasSarma,
  Tom Henighan, Ben Mann, Amanda Askell, Yuntao Bai, Anna Chen, et~al.
\newblock In-context learning and induction heads.
\newblock {\em arXiv preprint arXiv:2209.11895}, 2022.

\bibitem{arjovsky2016unitary}
Martin Arjovsky, Amar Shah, and Yoshua Bengio.
\newblock Unitary evolution recurrent neural networks.
\newblock In {\em International conference on machine learning}, pages
  1120--1128. PMLR, 2016.

\bibitem{juven2020cross}
Alexis Juven and Xavier Hinaut.
\newblock Cross-situational learning with reservoir computing for language
  acquisition modelling.
\newblock In {\em 2020 International Joint Conference on Neural Networks
  (IJCNN)}, pages 1--8. IEEE, 2020.

\bibitem{variengien2020journey}
Alexandre Variengien and Xavier Hinaut.
\newblock A journey in {ESN} and {LSTM} visualisations on a language task.
\newblock {\em arXiv preprint arXiv:2012.01748}, 2020.

\bibitem{paszke2019pytorch}
Adam Paszke, Sam Gross, Francisco Massa, Adam Lerer, James Bradbury, Gregory
  Chanan, Trevor Killeen, Zeming Lin, Natalia Gimelshein, Luca Antiga, et~al.
\newblock Pytorch: An imperative style, high-performance deep learning library.
\newblock {\em Advances in neural information processing systems}, 32, 2019.

\bibitem{kingma2014adam}
Diederik~P Kingma and Jimmy Ba.
\newblock Adam: A method for stochastic optimization.
\newblock {\em arXiv preprint arXiv:1412.6980}, 2014.

\bibitem{kaplan2020scaling}
Jared Kaplan, Sam McCandlish, Tom Henighan, Tom~B Brown, Benjamin Chess, Rewon
  Child, Scott Gray, Alec Radford, Jeffrey Wu, and Dario Amodei.
\newblock Scaling laws for neural language models.
\newblock {\em arXiv preprint arXiv:2001.08361}, 2020.

\bibitem{hoffmann2022training}
Jordan Hoffmann, Sebastian Borgeaud, Arthur Mensch, Elena Buchatskaya, Trevor
  Cai, Eliza Rutherford, DDL Casas, Lisa~Anne Hendricks, Johannes Welbl, Aidan
  Clark, et~al.
\newblock Training compute-optimal large language models.
\newblock {\em arXiv preprint arXiv:2203.15556}, 10, 2022.

\bibitem{bender2021dangers}
Emily~M Bender, Timnit Gebru, Angelina McMillan-Major, and Shmargaret
  Shmitchell.
\newblock On the dangers of stochastic parrots: Can language models be too big?
\newblock In {\em Proceedings of the 2021 ACM conference on fairness,
  accountability, and transparency}, pages 610--623, 2021.

\bibitem{chen2024role}
Lihu Chen and Ga{\"e}l Varoquaux.
\newblock What is the role of small models in the llm era: A survey.
\newblock {\em arXiv preprint arXiv:2409.06857}, 2024.

\end{thebibliography}


\newpage
\appendix

\section{Detailed Experimental Results}
\label{app:detailed_results}

\subsection{Best Overall (lower is better)}
\label{app:results_best}

\begin{table}[H]
\caption{Best overall experimental results for small (SM) and medium (MD) tasks. Values represent the single best performance achieved by each model across all evaluated parameter counts (1k, 10k, 100k) and learning rates (150 runs). Bold text indicates the top performance for a given task difficulty.}
\label{tab:result_best}
\makebox[\textwidth][c]{
    \begin{tabular}{lccccccc}
    \toprule
    Tâches & ESN & GRU & LSTM & MAMBA & TRANS (DO) & TRANS (ED) & XLSTM \\
    \midrule
    adding problem-SM                 & \textbf{0.00}  & \textbf{0.00}  & 0.02      & 0.01      & \textbf{0.00}               & \textbf{0.00}              & \textbf{0.00}  \\
    adding problem-MD                 & 0.69      & \textbf{0.00}  & \textbf{0.00}  & \textbf{0.00}  & \textbf{0.00}               & \textbf{0.00}              & \textbf{0.00}  \\
    associative  rec.-SM             & 0.44      & \textbf{0.35}  & 0.57      & 0.59      & 0.36                   & 0.39                  & 0.40      \\
    associative  rec.-MD             & 0.76      & 0.88      & 0.91      & 0.91      & \textbf{0.67}               & 0.78                  & 0.85      \\
    bracket matching-SM               & 0.13      & 0.18      & 0.21      & 0.29      & 0.19                   & 0.27                  & \textbf{0.06}  \\
    bracket matching-MD               & 0.13      & 0.01      & \textbf{0.00}  & 0.02      & 0.04                   & 0.01                  & \textbf{0.00}  \\
    chaotic forecast-SM            & \textbf{0.00}  & \textbf{0.00}  & 0.01      & \textbf{0.00}  & 0.04                   & 0.04                  & 0.01      \\
    chaotic forecast-MD            & \textbf{0.00}  & \textbf{0.00}  & \textbf{0.00}  & \textbf{0.00}  & 0.08                   & 0.07                  & \textbf{0.00}  \\
    c. pattern completion-SM  & 0.02      & \textbf{0.01}  & \textbf{0.01}  & \textbf{0.01}  & 0.03                   & 0.06                  & \textbf{0.01}  \\
    c. pattern completion-MD  & 0.06      & 0.01      & 0.01      & 0.02      & 0.01                   & \textbf{0.00}              & 0.01      \\
    c. postcasting-SM         & \textbf{0.00}  & \textbf{0.00}  & \textbf{0.00}  & \textbf{0.00}  & \textbf{0.00}               & \textbf{0.00}              & \textbf{0.00}  \\
    c. postcasting-MD         & \textbf{0.00}  & \textbf{0.00}  & \textbf{0.00}  & 0.02      & \textbf{0.00}               & \textbf{0.00}              & \textbf{0.00}  \\
    cross situation-SM                & \textbf{0.00}  & 0.03      & 0.05      & 0.05      & 0.03                   & 0.04                  & 0.03      \\
    cross situation-MD                & \textbf{0.00}  & \textbf{0.00}  & \textbf{0.00}  & \textbf{0.00}  & \textbf{0.00}               & \textbf{0.00}              & \textbf{0.00}  \\
    d. pattern completion-SM    & 0.06      & 0.06      & \textbf{0.05}  & 0.06      & 0.07                   & 0.36                  & 0.06      \\
    d. pattern completion-MD    & 0.09      & 0.18      & 0.11      & 0.14      & 0.07                   & \textbf{0.00}              & 0.18      \\
    d. postcasting-SM           & \textbf{0.00}  & \textbf{0.00}  & \textbf{0.00}  & \textbf{0.00}  & \textbf{0.00}               & \textbf{0.00}              & \textbf{0.00}  \\
    d. postcasting-MD           & \textbf{0.00}  & \textbf{0.00}  & \textbf{0.00}  & 0.08      & \textbf{0.00}               & \textbf{0.00}              & \textbf{0.00}  \\
    induction heads-SM                & 0.01      & 0.35      & 0.37      & 0.48      & \textbf{0.00}               & \textbf{0.00}              & 0.39      \\
    induction heads-MD                & 0.64      & 0.47      & 0.77      & 0.12      & \textbf{0.00}               & \textbf{0.00}              & 0.25      \\
    selective copy-SM                 & 0.41      & 0.22      & 0.52      & 0.21      & \textbf{0.05}               & 0.17                  & 0.34      \\
    selective copy-MD                 & 0.70      & 0.23      & 0.86      & \textbf{0.00}  & \textbf{0.00}               & 0.01                  & 0.23      \\
    simple copy-SM                    & \textbf{0.00}  & 0.49      & 0.57      & 0.42      & \textbf{0.00}               & \textbf{0.00}              & 0.42      \\
    simple copy-MD                    & 0.70      & 0.64      & 0.84      & 0.13      & \textbf{0.00}               & \textbf{0.00}              & 0.51      \\
    sinus forecast-SM              & \textbf{0.00}  & 0.01      & 0.01      & 0.01      & 0.02                   & 0.02                  & 0.01      \\
    sinus forecast-MD              & \textbf{0.00}  & 0.01      & \textbf{0.00}  & \textbf{0.00}  & 0.07                   & 0.07                  & 0.01      \\
    sorting problem-SM                & 0.43      & 0.42      & 0.49      & 0.51      & \textbf{0.00}               & \textbf{0.00}              & 0.05      \\
    sorting problem-MD                & 0.71      & 0.17      & 0.56      & \textbf{0.00}  & \textbf{0.00}               & \textbf{0.00}              & 0.06      \\
    \bottomrule
    \end{tabular}
}
\end{table}

\newpage
\subsection{Mean and Standard Deviation (lower is better)}
\label{app:results_mean}

\begin{table}[H]
\caption{Mean and Standard Deviation results (first half) for small (SM) and medium (MD) tasks. Performance is evaluated across varying parameter count (1k, 10k, 100k). Values represent the mean and standard deviation (mean ± std) of the best Learning Rate among 5, on 10 seeds (each value represents 50 runs). Bold text indicates the best performance achieved for a given task configuration. N/D indicate the lack of results.}
\label{tab:results_mean_1}
\makebox[\textwidth][c]{
    \begin{tabular}{lccccccc}
    \toprule
    Tâches & ESN & GRU & LSTM & MAMBA & TRANS. (DO) & TRANS. (ED) & XLSTM \\
    \midrule
    adding problem-SM-1k                   & 0.47±0.09      & \textbf{0.08±0.05}  & 0.50±0.31      & 0.63±0.21      & 0.56±0.24            & 0.69±0.06           & 0.32±0.37      \\
    adding problem-SM-10k                  & \textbf{0.02±0.03}  & 0.13±0.13      & 0.36±0.33      & 0.63±0.20      & 0.15±0.22            & 0.42±0.27           & 0.07±0.06      \\
    adding problem-SM-100k                 & N/D              & 0.16±0.10      & 0.55±0.29      & 0.61±0.23      & \textbf{0.04±0.03}        & 0.22±0.32           & 0.06±0.07      \\
    adding problem-MD-1k                   & 0.85±0.02      & \textbf{0.01±0.01}  & 0.02±0.01      & 0.88±0.01      & 0.70±0.29            & 0.88±0.01           & 0.44±0.45      \\
    adding problem-MD-10k                  & 0.71±0.02      & 0.01±0.00      & 0.01±0.01      & 0.23±0.37      & 0.01±0.01            & 0.54±0.44           & \textbf{0.00±0.00}  \\
    adding problem-MD-100k                 & N/D              & 0.01±0.00      & 0.02±0.01      & 0.23±0.38      & 0.01±0.01            & 0.20±0.36           & \textbf{0.00±0.00}  \\
    associative rec.-SM-1k               & \textbf{0.57±0.07}  & 0.66±0.03      & 0.66±0.04      & 0.65±0.04      & 0.64±0.04            & 0.65±0.05           & 0.64±0.05      \\
    associative rec.-SM-10k              & \textbf{0.52±0.04}  & 0.58±0.11      & 0.65±0.04      & 0.66±0.04      & 0.59±0.12            & 0.61±0.07           & 0.56±0.05      \\
    associative rec.-SM-100k             & N/D              & 0.57±0.11      & 0.65±0.07      & 0.66±0.04      & 0.55±0.09            & 0.56±0.10           & \textbf{0.52±0.07}  \\
    associative rec.-MD-1k               & \textbf{0.85±0.01}  & 0.93±0.01      & 0.93±0.01      & 0.93±0.01      & 0.92±0.01            & 0.93±0.01           & 0.92±0.01      \\
    associative rec.-MD-10k              & \textbf{0.78±0.01}  & 0.93±0.01      & 0.93±0.00      & 0.93±0.01      & 0.86±0.03            & 0.90±0.03           & 0.91±0.01      \\
    associative rec.-MD-100k             & N/D              & 0.90±0.01      & 0.93±0.01      & 0.93±0.01      & \textbf{0.84±0.07}        & 0.90±0.06           & 0.88±0.01      \\
    bracket matching-SM-1k                 & \textbf{0.25±0.11}  & 0.30±0.09      & 0.33±0.06      & 0.39±0.05      & 0.45±0.08            & 0.48±0.06           & 0.38±0.06      \\
    bracket matching-SM-10k                & \textbf{0.20±0.06}  & 0.35±0.03      & 0.36±0.06      & 0.38±0.04      & 0.46±0.05            & 0.46±0.06           & 0.28±0.10      \\
    bracket matching-SM-100k               & N/D              & 0.39±0.11      & 0.38±0.05      & 0.38±0.04      & 0.45±0.11            & 0.43±0.07           & \textbf{0.32±0.10}  \\
    bracket matching-MD-1k                 & 0.20±0.02      & \textbf{0.06±0.09}  & 0.08±0.11      & 0.27±0.11      & 0.42±0.10            & 0.47±0.05           & 0.17±0.13      \\
    bracket matching-MD-10k                & 0.16±0.02      & 0.14±0.13      & 0.07±0.11      & 0.19±0.11      & 0.23±0.14            & 0.37±0.14           & \textbf{0.03±0.01}  \\
    bracket matching-MD-100k               & N/D              & 0.11±0.11      & 0.20±0.16      & 0.19±0.13      & 0.11±0.07            & 0.14±0.14           & \textbf{0.02±0.01}  \\
    chaotic forecast-SM-1k              & \textbf{0.00±0.00}  & 0.04±0.02      & 0.04±0.01      & 0.04±0.02      & 0.11±0.04            & 0.08±0.01           & 0.06±0.02      \\
    chaotic forecast-SM-10k             & \textbf{0.00±0.00}  & 0.02±0.01      & 0.02±0.01      & 0.01±0.01      & 0.07±0.01            & 0.07±0.01           & 0.05±0.02      \\
    chaotic forecast-SM-100k            & N/D              & 0.02±0.02      & 0.03±0.01      & \textbf{0.01±0.01}  & 0.06±0.02            & 0.05±0.01           & 0.02±0.01      \\
    chaotic forecast-MD-1k              & \textbf{0.00±0.00}  & 0.02±0.03      & 0.01±0.02      & \textbf{0.00±0.00}  & 0.11±0.02            & 0.10±0.00           & 0.03±0.02      \\
    chaotic forecast-MD-10k             & \textbf{0.00±0.00}  & 0.01±0.02      & 0.02±0.01      & \textbf{0.00±0.00}  & 0.16±0.04            & 0.11±0.02           & 0.01±0.01      \\
    chaotic forecast-MD-100k            & N/D              & \textbf{0.00±0.01}  & 0.02±0.01      & \textbf{0.00±0.00}  & 0.10±0.02            & 0.09±0.01           & \textbf{0.00±0.00}  \\
    c. pattern comp.-SM-1k    & 0.03±0.00      & \textbf{0.01±0.00}  & 0.02±0.01      & 0.04±0.03      & 0.07±0.02            & 0.08±0.01           & 0.03±0.01      \\
    c. pattern comp.-SM-10k   & 0.04±0.01      & \textbf{0.01±0.00}  & \textbf{0.01±0.00}  & 0.02±0.00      & 0.07±0.01            & 0.07±0.01           & 0.02±0.00      \\
    c. pattern comp.-SM-100k  & N/D              & \textbf{0.01±0.00}  & \textbf{0.01±0.00}  & 0.02±0.00      & 0.07±0.01            & 0.07±0.01           & 0.02±0.01      \\
    c. pattern comp.-MD-1k    & 0.08±0.00      & \textbf{0.01±0.00}  & 0.02±0.00      & 0.04±0.02      & 0.07±0.02            & 0.08±0.00           & 0.05±0.00      \\
    c. pattern comp.-MD-10k   & 0.06±0.00      & \textbf{0.01±0.00}  & \textbf{0.01±0.00}  & 0.02±0.00      & \textbf{0.01±0.00}        & 0.07±0.02           & 0.03±0.01      \\
    c. pattern comp.-MD-100k  & N/D              & \textbf{0.01±0.00}  & \textbf{0.01±0.00}  & 0.02±0.00      & \textbf{0.01±0.00}        & 0.03±0.04           & 0.02±0.00      \\
    c. postcasting-SM-1k           & \textbf{0.00±0.00}  & \textbf{0.00±0.00}  & \textbf{0.00±0.00}  & \textbf{0.00±0.00}  & 0.04±0.09            & 0.15±0.08           & 0.06±0.01      \\
    c. postcasting-SM-10k          & \textbf{0.00±0.00}  & \textbf{0.00±0.00}  & \textbf{0.00±0.00}  & \textbf{0.00±0.00}  & \textbf{0.00±0.00}        & \textbf{0.00±0.00}       & \textbf{0.00±0.00}  \\
    c. postcasting-SM-100k         & N/D              & \textbf{0.00±0.00}  & \textbf{0.00±0.00}  & \textbf{0.00±0.00}  & \textbf{0.00±0.00}        & \textbf{0.00±0.00}       & \textbf{0.00±0.00}  \\
    c. postcasting-MD-1k           & 0.07±0.02      & \textbf{0.00±0.01}  & 0.01±0.01      & 0.15±0.04      & \textbf{0.00±0.00}        & 0.10±0.08           & 0.20±0.00      \\
    c. postcasting-MD-10k          & \textbf{0.00±0.00}  & \textbf{0.00±0.00}  & \textbf{0.00±0.00}  & 0.06±0.03      & \textbf{0.00±0.00}        & \textbf{0.00±0.00}       & 0.07±0.03      \\
    c. postcasting-MD-100k         & N/D              & \textbf{0.00±0.00}  & \textbf{0.00±0.00}  & 0.09±0.04      & \textbf{0.00±0.00}        & \textbf{0.00±0.00}       & \textbf{0.00±0.00}  \\
    cross situation-SM-1k                  & \textbf{0.01±0.01}  & 0.23±0.22      & 0.29±0.22      & 0.55±0.21      & 0.61±0.17            & 0.70±0.00           & 0.50±0.22      \\
    cross situation-SM-10k                 & \textbf{0.01±0.01}  & 0.07±0.02      & 0.08±0.03      & 0.16±0.06      & 0.09±0.03            & 0.22±0.25           & 0.12±0.04      \\
    cross situation-SM-100k                & N/D              & \textbf{0.04±0.01}  & 0.08±0.01      & 0.14±0.04      & 0.06±0.01            & 0.13±0.20           & 0.05±0.01      \\
    cross situation-MD-1k                  & 0.13±0.01      & \textbf{0.07±0.01}  & 0.10±0.01      & 0.56±0.18      & 0.52±0.27            & 0.80±0.00           & 0.57±0.20      \\
    cross situation-MD-10k                 & \textbf{0.00±0.00}  & \textbf{0.00±0.00}  & \textbf{0.00±0.00}  & \textbf{0.00±0.00}  & 0.01±0.00            & 0.04±0.01           & \textbf{0.00±0.00}  \\
    cross situation-MD-100k                & N/D              & \textbf{0.00±0.00}  & \textbf{0.00±0.00}  & \textbf{0.00±0.00}  & \textbf{0.00±0.00}        & \textbf{0.00±0.00}       & \textbf{0.00±0.00}  \\
    \bottomrule
    \end{tabular}
}
\end{table}

\begin{table}[H]
\caption{Complete experimental results (second half) for small (SM) and medium (MD) tasks. Performance is evaluated across varying parameter count (1k, 10k, 100k). Values represent the mean and standard deviation (mean ± std) of the best Learning Rate among 5, on 10 seeds (each value represents 50 runs). Bold text indicates the best performance achieved for a given task configuration. N/D indicate the lack of results.}
\label{tab:results_mean_2}
\makebox[\textwidth][c]{
\begin{tabular}{lccccccc}
\toprule
Tâches & ESN & GRU & LSTM & MAMBA & TRANS. (DO) & TRANS. (ED) & XLSTM \\
\midrule
d. pattern comp.-SM-1k      & \textbf{0.07±0.01}  & 0.17±0.04      & 0.24±0.15      & 0.17±0.12      & 0.41±0.09            & 0.46±0.09           & 0.13±0.02      \\
d. pattern comp.-SM-10k     & \textbf{0.07±0.01}  & 0.11±0.02      & 0.15±0.15      & 0.10±0.02      & 0.38±0.14            & 0.41±0.02           & 0.10±0.02      \\
d. pattern comp.-SM-100k    & N/D              & \textbf{0.07±0.03}  & 0.11±0.02      & 0.09±0.01      & 0.22±0.18            & 0.41±0.02           & 0.09±0.02      \\
d. pattern comp.-MD-1k      & 0.57±0.06      & 0.59±0.03      & 0.69±0.11      & 0.63±0.16      & 0.63±0.21            & 0.80±0.02           & \textbf{0.41±0.05}  \\
d. pattern comp.-MD-10k     & 0.10±0.00      & 0.27±0.03      & 0.25±0.10      & 0.20±0.03      & \textbf{0.08±0.00}        & 0.56±0.26           & 0.20±0.01      \\
d. pattern comp.-MD-100k    & N/D              & 0.29±0.02      & 0.30±0.03      & 0.19±0.03      & 0.08±0.01            & \textbf{0.00±0.00}       & 0.18±0.01      \\
d. postcasting-SM-1k             & \textbf{0.00±0.00}  & \textbf{0.00±0.00}  & \textbf{0.00±0.00}  & \textbf{0.00±0.00}  & 0.03±0.10            & 0.38±0.28           & 0.22±0.11      \\
d. postcasting-SM-10k            & \textbf{0.00±0.00}  & \textbf{0.00±0.00}  & \textbf{0.00±0.00}  & \textbf{0.00±0.00}  & \textbf{0.00±0.00}        & \textbf{0.00±0.00}       & \textbf{0.00±0.00}  \\
d. postcasting-SM-100k           & N/D              & \textbf{0.00±0.01}  & \textbf{0.00±0.00}  & \textbf{0.00±0.00}  & \textbf{0.00±0.00}        & \textbf{0.00±0.00}       & \textbf{0.00±0.00}  \\
d. postcasting-MD-1k             & 0.63±0.01      & 0.64±0.04      & 0.65±0.09      & 0.74±0.07      & \textbf{0.00±0.00}        & 0.67±0.20           & 0.83±0.01      \\
d. postcasting-MD-10k            & \textbf{0.00±0.00}  & \textbf{0.00±0.00}  & \textbf{0.00±0.01}  & 0.30±0.06      & \textbf{0.00±0.00}        & \textbf{0.00±0.00}       & 0.54±0.03      \\
d. postcasting-MD-100k           & N/D              & \textbf{0.00±0.00}  & \textbf{0.00±0.00}  & 0.12±0.03      & \textbf{0.00±0.00}        & \textbf{0.00±0.00}       & 0.01±0.01      \\
induction heads-SM-1k                  & \textbf{0.36±0.02}  & 0.60±0.09      & 0.62±0.08      & 0.65±0.04      & 0.40±0.34            & 0.64±0.04           & 0.53±0.09      \\
induction heads-SM-10k                 & 0.02±0.00      & 0.53±0.12      & 0.59±0.09      & 0.60±0.07      & \textbf{0.00±0.00}        & 0.18±0.29           & 0.46±0.07      \\
induction heads-SM-100k                & N/D              & 0.49±0.13      & 0.48±0.08      & 0.58±0.04      & \textbf{0.00±0.00}        & \textbf{0.00±0.00}       & 0.44±0.03      \\
induction heads-MD-1k                  & 0.73±0.00      & 0.81±0.06      & 0.87±0.00      & 0.84±0.06      & \textbf{0.35±0.31}        & 0.77±0.05           & 0.74±0.02      \\
induction heads-MD-10k                 & 0.64±0.00      & 0.58±0.11      & 0.87±0.00      & 0.31±0.06      & \textbf{0.00±0.00}        & \textbf{0.00±0.00}       & 0.39±0.07      \\
induction heads-MD-100k                & N/D              & 0.60±0.13      & 0.85±0.04      & 0.21±0.05      & \textbf{0.00±0.00}        & \textbf{0.00±0.00}       & 0.29±0.01 \\
selective copy-SM-1k                   & \textbf{0.48±0.04}  & 0.63±0.08      & 0.66±0.02      & 0.63±0.05      & 0.59±0.11            & 0.66±0.02           & 0.51±0.08      \\
selective copy-SM-10k                  & 0.44±0.02      & 0.54±0.16      & 0.65±0.02      & 0.55±0.13      & \textbf{0.37±0.27}        & 0.63±0.06           & 0.44±0.04      \\
selective copy-SM-100k                 & N/D              & 0.61±0.11      & 0.65±0.01      & 0.52±0.14      & \textbf{0.10±0.05}        & 0.43±0.21           & 0.42±0.04      \\
selective copy-MD-1k                   & 0.75±0.01      & 0.65±0.13      & 0.87±0.00      & 0.78±0.14      & \textbf{0.58±0.18}        & 0.87±0.00           & 0.76±0.02      \\
selective copy-MD-10k                  & 0.70±0.00      & 0.54±0.25      & 0.87±0.00      & 0.32±0.35      & \textbf{0.04±0.07}        & 0.20±0.28           & 0.45±0.08      \\
selective copy-MD-100k                 & N/D              & 0.54±0.24      & 0.88±0.00      & 0.11±0.16      & \textbf{0.01±0.00}        & 0.02±0.03           & 0.29±0.06      \\
simple copy-SM-1k                      & 0.47±0.01      & 0.64±0.06      & 0.67±0.01      & 0.65±0.04      & \textbf{0.26±0.33}        & 0.65±0.03           & 0.53±0.05      \\
simple copy-SM-10k                     & \textbf{0.00±0.00}  & 0.62±0.06      & 0.66±0.02      & 0.61±0.05      & \textbf{0.00±0.00}        & 0.12±0.26           & 0.49±0.03      \\
simple copy-SM-100k                    & N/D              & 0.63±0.05      & 0.65±0.03      & 0.50±0.05      & \textbf{0.00±0.00}        & \textbf{0.00±0.00}       & 0.46±0.02      \\
simple copy-MD-1k                      & 0.76±0.00      & 0.82±0.05      & 0.87±0.00      & 0.81±0.02      & \textbf{0.04±0.09}        & 0.83±0.05           & 0.78±0.00      \\
simple copy-MD-10k                     & 0.70±0.00      & 0.72±0.03      & 0.87±0.00      & 0.50±0.08      & \textbf{0.00±0.00}        & \textbf{0.00±0.00}       & 0.69±0.03      \\
simple copy-MD-100k                    & N/D              & 0.74±0.08      & 0.87±0.00      & 0.25±0.05      & \textbf{0.00±0.00}        & \textbf{0.00±0.00}       & 0.55±0.03      \\
sinus forecast-SM-1k                & \textbf{0.00±0.00}  & 0.01±0.00      & 0.02±0.00      & 0.02±0.01      & 0.16±0.12            & 0.31±0.21           & 0.01±0.00      \\
sinus forecast-SM-10k               & \textbf{0.00±0.00}  & 0.01±0.00      & 0.03±0.01      & 0.01±0.00      & 0.07±0.02            & 0.12±0.05           & 0.01±0.00      \\
sinus forecast-SM-100k              & N/D              & 0.02±0.00      & 0.05±0.01      & \textbf{0.01±0.01}  & 0.05±0.05            & 0.06±0.03           & \textbf{0.01±0.01}  \\
sinus forecast-MD-1k                & \textbf{0.00±0.00}  & 0.05±0.01      & 0.03±0.01      & 0.04±0.02      & 0.35±0.27            & 0.18±0.12           & 0.05±0.01      \\
sinus forecast-MD-10k               & \textbf{0.00±0.00}  & 0.03±0.01      & 0.03±0.02      & 0.03±0.02      & 0.10±0.01            & 0.20±0.09           & 0.04±0.02      \\
sinus forecast-MD-100k              & N/D              & 0.03±0.02      & 0.04±0.02      & \textbf{0.02±0.02}  & 0.08±0.01            & 0.10±0.02           & 0.04±0.02      \\
sorting problem-SM-1k                  & 0.51±0.01      & 0.49±0.03      & 0.51±0.01      & 0.61±0.05      & \textbf{0.08±0.21}        & 0.67±0.02           & 0.54±0.05      \\
sorting problem-SM-10k                 & 0.46±0.01      & 0.50±0.02      & 0.52±0.03      & 0.55±0.03      & \textbf{0.00±0.00}        & 0.06±0.16           & 0.31±0.02      \\
sorting problem-SM-100k                & N/D              & 0.51±0.01      & 0.52±0.05      & 0.53±0.01      & \textbf{0.00±0.00}        & 0.05±0.16           & 0.22±0.08      \\
sorting problem-MD-1k                  & 0.75±0.00      & 0.75±0.01      & 0.77±0.01      & 0.70±0.07      & \textbf{0.00±0.00}        & 0.87±0.00           & 0.70±0.04      \\
sorting problem-MD-10k                 & 0.72±0.00      & 0.33±0.04      & 0.61±0.03      & 0.20±0.13      & \textbf{0.00±0.00}        & \textbf{0.00±0.00}       & 0.28±0.08      \\
sorting problem-MD-100k                & N/D              & 0.24±0.05      & 0.70±0.01      & 0.02±0.02      & \textbf{0.00±0.00}        & \textbf{0.00±0.00}       & 0.11±0.04      \\
\bottomrule
\end{tabular}
}
\end{table}

\newpage
\newpage

\section{Task Configuration: Small \& Medium}
\label{app:task_config}

This section details the specific parameters used to generate the datasets for both the \texttt{small (SM)} and \texttt{medium (MD)} difficulty across all tasks. 

\begin{itemize}

    \item \textbf{Sinus Forecasting}
    \begin{itemize}
        \item \textbf{Small:}\\\texttt{sequence\_length=200, forecast\_length=5, training\_ratio=0.45, validation\_ratio=0.1, testing\_ratio=0.45}
        \item \textbf{Medium:}\\ \texttt{sequence\_length=2000, forecast\_length=15, training\_ratio=0.45, validation\_ratio=0.1, testing\_ratio=0.45}
    \end{itemize}

    \item \textbf{Chaotic Forecasting}
    \begin{itemize}
        \item \textbf{Small:}\\ \texttt{sequence\_length=200, forecast\_length=5, training\_ratio=0.45, validation\_ratio=0.1, testing\_ratio=0.45}
        \item \textbf{Medium:}\\ \texttt{sequence\_length=2000, forecast\_length=15, training\_ratio=0.45, validation\_ratio=0.1, testing\_ratio=0.45}
    \end{itemize}

    \item \textbf{Discrete Postcasting}
    \begin{itemize}
        \item \textbf{Small:}\\ \texttt{n\_train=100, n\_valid=20, n\_test=100, sequence\_length=50, delay=5, n\_symbols=3}
        \item \textbf{Medium:}\\ \texttt{n\_train=1000, n\_valid=200, n\_test=1000, sequence\_length=100, delay=15, n\_symbols=8}
    \end{itemize}

    \item \textbf{Continuous Postcasting}
    \begin{itemize}
        \item \textbf{Small:}\\ \texttt{n\_train=100, n\_valid=20, n\_test=100, sequence\_length=50, delay=5}
        \item \textbf{Medium:}\\ \texttt{n\_train=1000, n\_valid=200, n\_test=1000, sequence\_length=100, delay=15}
    \end{itemize}

    \item \textbf{Discrete Pattern Completion}
    \begin{itemize}
        \item \textbf{Small:}\\ \texttt{n\_train=100, n\_valid=20, n\_test=100, sequence\_length=60, n\_symbols=3, base\_length=4, mask\_ratio=0.2}
        \item \textbf{Medium:}\\ \texttt{n\_train=1000, n\_valid=200, n\_test=1000, sequence\_length=150, n\_symbols=8, base\_length=10, mask\_ratio=0.2}
    \end{itemize}

    \item \textbf{Continuous Pattern Completion}
    \begin{itemize}
        \item \textbf{Small:}\\ \texttt{n\_train=100, n\_valid=20, n\_test=100, sequence\_length=60, base\_length=4, mask\_ratio=0.2}
        \item \textbf{Medium:}\\ \texttt{n\_train=1000, n\_valid=200, n\_test=1000, sequence\_length=150, base\_length=10, mask\_ratio=0.2}
    \end{itemize}

    \item \textbf{Bracket Matching}
    \begin{itemize}
        \item \textbf{Small:}\\ \texttt{n\_train=100, n\_valid=20, n\_test=100, sequence\_length=50, max\_depth=5}
        \item \textbf{Medium:}\\ \texttt{n\_train=1000, n\_valid=200, n\_test=1000, sequence\_length=100, max\_depth=10}
    \end{itemize}

    \item \textbf{Simple Copy}
    \begin{itemize}
        \item \textbf{Small:}\\ \texttt{n\_train=100, n\_valid=20, n\_test=100, sequence\_length=22, delay=5, n\_symbols=3}
        \item \textbf{Medium:}\\ \texttt{n\_train=1000, n\_valid=200, n\_test=1000, sequence\_length=50, delay=10, n\_symbols=8}
    \end{itemize}

    \item \textbf{Selective Copy}
    \begin{itemize}
        \item \textbf{Small:}\\ \texttt{n\_train=100, n\_valid=20, n\_test=100, sequence\_length=40, delay=5, n\_markers=5, n\_symbols=3}
        \item \textbf{Medium:}\\ \texttt{n\_train=1000, n\_valid=200, n\_test=1000, sequence\_length=80, delay=10, n\_markers=10, n\_symbols=8}
    \end{itemize}

    \item \textbf{Adding Problem}
    \begin{itemize}
        \item \textbf{Small:}\\ \texttt{n\_train=100, n\_valid=20, n\_test=100, sequence\_length=10, max\_number=3}
        \item \textbf{Medium:}\\ \texttt{n\_train=1000, n\_valid=200, n\_test=1000, sequence\_length=20, max\_number=8}
    \end{itemize}

    \item \textbf{Sorting Problem}
    \begin{itemize}
        \item \textbf{Small:}\\ \texttt{n\_train=100, n\_valid=20, n\_test=100, sequence\_length=10, n\_symbols=3}
        \item \textbf{Medium:}\\ \texttt{n\_train=1000, n\_valid=200, n\_test=1000, sequence\_length=20, n\_symbols=8}
    \end{itemize}

    \item \textbf{Cross Situation}
    \begin{itemize}
        \item \textbf{Small:}\\ \texttt{n\_train=100, n\_valid=20, n\_test=100, objects=['glass', 'orange'], colors=['blue', 'orange'], positions=['left', 'right']}
        \item \textbf{Medium:}\\ \texttt{n\_train=1000, n\_valid=200, n\_test=1000, objects=['glass', 'orange', 'cup', 'bowl'], colors=['blue', 'orange', 'green', 'red'], positions=['left', 'right', ('center', 'middle')]}
    \end{itemize}

    \item \textbf{Associative Recall}
    \begin{itemize}
        \item \textbf{Small:}\\ \texttt{n\_train=100, n\_valid=20, n\_test=100, sequence\_length=16, num\_pairs=3, n\_symbols=5}
        \item \textbf{Medium:}\\ \texttt{n\_train=1000, n\_valid=200, n\_test=1000, sequence\_length=32, num\_pairs=7, n\_symbols=16}
    \end{itemize}

    \item \textbf{Induction Heads}
    \begin{itemize}
        \item \textbf{Small:}\\ \texttt{n\_train=100, n\_valid=20, n\_test=100, sequence\_length=40, n\_symbols=3}
        \item \textbf{Medium:}\\ \texttt{n\_train=1000, n\_valid=200, n\_test=1000, sequence\_length=100, n\_symbols=8}
    \end{itemize}

\end{itemize}


\end{document}